# Differential privacy enables fair and accurate AI-based analysis of speech disorders while protecting patient data

Soroosh Tayebi Arasteh (1,2,3,4), Mahshad Lotfinia (2), Paula Andrea Perez-Toro (1,5), Tomas Arias-Vergara (1,5), Mahtab Ranji (1), Juan Rafael Orozco-Arroyave (1,5), Maria Schuster (6), Andreas Maier (1), Seung Hee Yang (7)

(1) Pattern Recognition Lab, Friedrich-Alexander-Universität Erlangen-Nürnberg, Erlangen, Germany.
(2) Department of Diagnostic and Interventional Radiology, University Hospital RWTH Aachen, Aachen, Germany.
(3) Department of Urology, Stanford University, Stanford, CA, USA.
(4) Department of Radiology, Stanford University, Stanford, CA, USA.
(5) GITA Lab, Faculty of Engineering, University of Antioquia, Medellín, Colombia.
(6) Department of Otorhinolaryngology, Head and Neck Surgery, Ludwig-Maximilians-Universität München, Munich, Germany.
(7) Speech & Language Processing Lab, Friedrich-Alexander-Universität Erlangen-Nürnberg, Erlangen, Germany.

## Abstract

Speech pathology has impacts on communication abilities and quality of life. While deep learning-based models have shown potential in diagnosing these disorders, the use of sensitive data raises critical privacy concerns. Although differential privacy (DP) has been explored in the medical imaging domain, its application in pathological speech analysis remains largely unexplored despite the equally critical privacy concerns. To the best of our knowledge, this study is the first to investigate DP's impact on pathological speech data, focusing on the trade-offs between privacy, diagnostic accuracy, and fairness. Using a large, real-world dataset of 200 hours of recordings from 2,839 German-speaking participants, we observed a maximum accuracy reduction of 3.85% when training with DP with high privacy levels. To highlight real-world privacy risks, we demonstrated the vulnerability of non-private models to gradient inversion attacks, reconstructing identifiable speech samples and showcasing DP's effectiveness in mitigating these risks. To explore the potential generalizability across languages and disorders, we validated our approach on a dataset of Spanish-speaking Parkinson's disease patients, leveraging pretrained models from healthy English-speaking datasets, and demonstrated that careful pretraining on large-scale task-specific datasets can maintain favorable accuracy under DP constraints. A comprehensive fairness analysis revealed minimal gender bias at reasonable privacy levels but underscored the need for addressing age-related disparities. Our results establish that DP can balance privacy and utility in speech disorder detection, while highlighting unique challenges in privacy-fairness trade-offs for speech data. This provides a foundation for refining DP methodologies and improving fairness across diverse patient groups in real-world deployments.

**Correspondence**

Soroosh Tayebi Arasteh, PhD, PhD
Pattern Recognition Lab
Friedrich-Alexander-Universität Erlangen-Nürnberg
Martensstr. 3
91058 Erlangen, Germany
Email: [soroosh.arasteh@fau.de](soroosh.arasteh@fau.de)

# 1. Introduction

Speech pathology, which refers to speech impairments caused by various disorders, is a critical area of study due to its important impact on an individual's quality of life and communication abilities[1,2]. Early and accurate detection of speech disorders can lead to more effective interventions and improved outcomes for patients. Artificial intelligence (AI)-based models have shown remarkable potential in diagnosing and analyzing these speech disorders by leveraging vast amounts of data to identify patterns that may not be apparent to human clinicians[3–5]. Studies have highlighted the expanding role of pathological speech in evaluating neurological conditions such as Parkinson's[3,6] and Alzheimer's[4], as well as speech disorders like Dysarthria and Dysglossia[7,8]. However, the integration of AI in this sensitive field raises substantial concerns about patient privacy[9–12]. Recent research[9] has shown that pathological speech, as a biomarker, is more vulnerable to re-identification attacks compared to healthy speech, making the protection of patient data confidentiality crucial. Misuse or unauthorized access to such data can result in severe ethical and legal consequences.

In response, several privacy-preserving strategies have been explored. Federated learning (FL) enables decentralized training without requiring raw data sharing[13–15], but its reliance on model parameter exchange leaves room for privacy leakage through adversarial attacks[16–18]. Similarly, anonymization[10–12] methods attempt to obscure speaker identity prior to model training[19]. While these approaches can reduce privacy risks, they are not pathology-agnostic and may unintentionally distort clinically relevant information. Moreover, anonymized or federated data can remain susceptible to re-identification[17,19], underscoring the need for more robust and formally grounded privacy mechanisms in AI-driven speech disorder detection[20,21].

These challenges underscore the need for more robust privacy-preserving techniques in AI-based speech disorder detection, leading to the motivation for adopting differential privacy (DP)[22]. Unlike traditional methods, DP provides a formal and quantifiable framework for protecting sensitive information, even in FL or other distributed training environments where privacy risks are elevated[16,22]. In such settings, adversaries can exploit vulnerabilities to extract detailed information during the training process or manipulate the model itself, posing major threats to patient privacy. Models trained on sensitive medical data, including pathological speech, are particularly vulnerable to attacks like membership inference and model inversion, where attackers can reconstruct aspects of the original training data[16,21,23,24]. This risk is heightened in scenarios with smaller datasets, a common issue in medical AI due to data scarcity. DP addresses these concerns by limiting the amount of information that any single data point can contribute to the model, offering a robust defense against re-identification and other privacy threats[21,25]. Furthermore, DP not only provides formal privacy guarantees but has also been empirically shown to mitigate the risks associated with membership inference and data reconstruction attacks. By controlling the privacy budget, DP allows for a balance between maintaining privacy and preserving the utility of the data, though it is important to note that absolute privacy—where no risk exists—is only achievable if no information is shared, as seen in encryption methods[26]. While encryption ensures perfect privacy as long as data remains encrypted[17], DP offers a practical solution for situations where data must be used, such as in model predictions, by providing a provable safeguard against sophisticated adversarial attacks, thereby aligning with modern privacy standards[16].

Determining the appropriate privacy budget in DP is a major challenge, as it requires a careful balance between privacy protection and model utility[16]. While it is technically possible to assess the risk of successful attacks relative to model utility at a given privacy budget, these trade-offs extend beyond technical considerations. They also involve ethical, societal, and political factors, particularly in sensitive fields like medical AI. One major trade-off is the privacy-utility trade-off, where stronger privacy guarantees may lead to a reduction in diagnostic accuracy—a critical concern in medical applications where precise diagnoses are vital[27,28]. Additionally, there is a trade-off between privacy and fairness[16,21,24]. DP can unintentionally exacerbate demographic disparities in AI models by limiting the information learned about under-represented patient groups in the training data, potentially leading to biased predictions or diagnoses[29,30]. In healthcare, where fairness and equity are crucial, managing these trade-offs is essential to effectively applying DP in AI-based speech disorder detection without compromising diagnostic accuracy or fairness across different patient groups[16].

Most prior work on DP has focused on image-based approaches[31,32], such as those found in medical imaging. For example, some studies have investigated privacy-utility trade-offs in FL schemes combined with DP methods on brain tumor segmentation datasets[33] and chest X-ray classification[34]. One study demonstrated that while DP training for chest X-ray classification results in slightly lower accuracy, it does not substantially increase discrimination based on age, sex, or co-morbidity[16]. Another study further showed that the cross-institutional performance of these models remained stable under DP, with negligible trade-offs in accuracy[24]. Despite these advances, the application of DP in pathological speech remains largely unexplored, even though privacy concerns in this domain are as critical as in medical imaging[35–38]. Furthermore, most prior work with speech data has focused on healthy speech[39] or only considered accuracy implications of DP[40], leaving a gap in understanding its broader impact on pathological speech data.

Given the critical importance of protecting sensitive patient information in speech disorder detection, our study undertakes the first comprehensive investigation into the application of DP in the context of pathological speech data. This research explores the use of DP in training complex diagnostic AI models on a large-scale, real-world, multi-institutional pathological speech dataset, providing an extensive evaluation of both privacy-utility and privacy-fairness trade-offs. Our work aims to fill a substantial gap in the literature and offer a foundational understanding of how DP can be effectively implemented in pathological speech analysis. This research is particularly relevant to healthcare providers, AI researchers in medicine, and regulatory authorities, including legislative bodies, institutional review boards, and data protection officers[16,41]. We have meticulously designed our study to address the most pressing concerns in this area, focusing on rigorous assessments of diagnostic accuracy, robust privacy safeguards, and the equitable treatment of diverse patient groups. By providing these critical insights, we aim to support the development of AI models for pathological speech analysis that are not only effective but also ethically and legally sound, ensuring their safe and fair application in real-world medical speech environments.

In this study, we conduct a detailed investigation into how DP affects the diagnostic performance of models trained on pathological speech data (see **Figure 1**). To the best of our knowledge, this is the first study of its scale to analyze patient privacy considerations in pathological speech data and the subsequent utility and fairness trade-offs. Our main contributions can be summarized as follows: (i) We analyze the diagnostic accuracy reductions imposed by DP training of DL models using a large, real-world dataset[9,19,42] consisting of approximately 200 hours of recordings from n=2,839 German-speaking participants, which includes both pathological and healthy speech samples. The dataset covers speech disorders such as Dysarthria and Dysglossia, pathological conditions like Cleft Lip and Palate (CLP), as well as healthy controls. We document a maximum accuracy reduction of only 3.85% when utilizing the Differentially Private Stochastic Gradient Descent (DP-SGD) algorithm[43] in training the diagnostic DL model with privacy budgets of $\varepsilon = 7.51$ and $\delta = 0.001$, while effectively ensuring patient privacy protection. (ii) To assess the potential generalizability of our findings, we explore a second pathological dataset[44], considering three axes of variation: (a) the task of Parkinson's disease (PD) detection, a neurological disorder, (b) using data from Spanish-speaking patients, and (c) applying a much smaller dataset (n=100) participants). Validating previous results from medical imaging datasets[16], we demonstrate that careful pretraining on large-scale task-specific pathological datasets using DP-compatible convolutional neural networks (CNNs) can result in private models achieving accuracy on par with, or even slightly better than, non-private models. (iii) We perform a comprehensive analysis of fairness bias under privacy constraints across different demographic groups. We find that, as long as extremely high privacy levels ($\varepsilon \approx 1$)—which are not commonly required in practice—are avoided, privacy constraints within a more realistic range ($2 < \varepsilon < 10$) do not introduce substantial discrimination between female and male patients. However, greater attention should be given to ensuring equity across different age groups to avoid unfair biases.

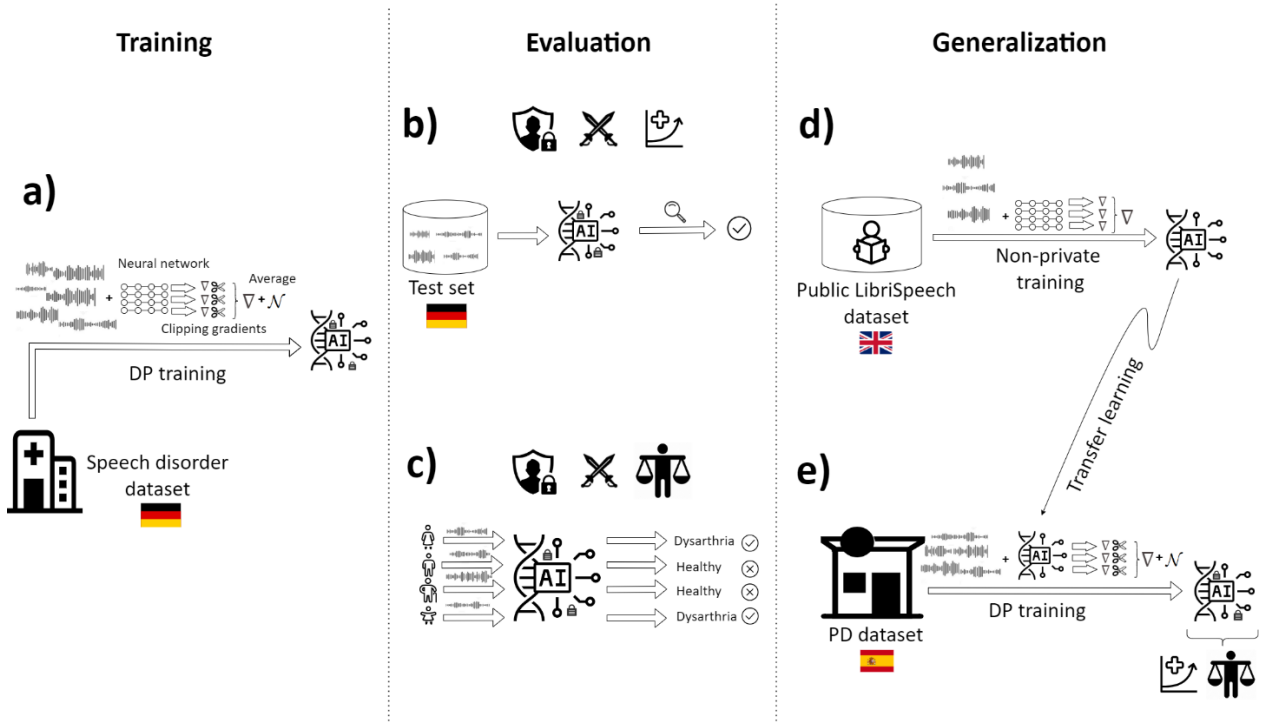

**Figure 1: Overview of the methodology. (a)** Differential privacy (DP) is applied to train an AI model on a large German speech disorder dataset (n=2,839) for diagnosing Dysarthria, Dysglossia, and detecting Cleft Lip and Palate as well as healthy participants, ensuring mathematical privacy guarantees. **(b)** The privacy-utility trade-off is evaluated using the held-out test set. **(c)** The privacy-fairness trade-off is assessed across demographic subgroups (e.g., sex, age) to identify potential biases introduced by DP training compared to non-private training. **(d)** To generalize results across different languages, tasks, and data sizes, the large-scale LibriSpeech[46] dataset of healthy English speakers is used for non-private pretraining on general speech features. **(e)** The pretrained model is then used as weight initialization for DP training on the smaller Spanish PC-GITA dataset[44] (n=100) for Parkinson's Disease (PD) detection, addressing the challenge of DP training with under-represented data and aiming to develop an accurate and fair PD detection model.

# 2. Results

## 2.1. High diagnostic performance under privacy constraints

We first evaluated the diagnostic performance of models trained with DP compared to non-private models. Using a large-scale, multi-institutional pathological speech dataset of German speakers[9,19,42], which includes n=1,979 training speakers and n=860 held-out test speakers (see **Table 1** for dataset characteristics), we addressed the multiclass detection of speech disorders and pathological conditions. Specifically, the tasks involved detecting speech disorders such as Dysarthria and Dysglossia, identifying the pathological condition CLP, and distinguishing healthy speech. Two distinct neural networks were trained on the training set: one using DP and the other without DP. Both models were then tested on the same held-out test set to evaluate their performance. **Figure 2** presents the evaluation results for both non-private and DP-trained models across different ε values.

For the non-private model, the average area under the receiver operating characteristic curve (AUROC) was 99.92 ± 0.02% [95% CI: 99.90, 99.93], with an accuracy of 99.10 ± 0.24% [95% CI: 98.96, 99.24] across all disorders, conditions, and control groups. Specifically, the AUROC values were 99.90 ± 0.02% [95% CI: 99.86, 99.93] for Dysarthria, 99.94 ± 0.01% [95% CI: 99.93, 99.96] for Dysglossia, 99.91 ± 0.03% [95% CI: 99.86, 99.95] for CLP, and 99.91 ± 0.01% [95% CI: 99.89, 99.93] for the control group. Corresponding accuracy values were 99.08 ± 0.18% [95% CI: 98.75, 99.42] for Dysarthria, 98.99 ± 0.14% [95% CI: 98.69, 99.19] for Dysglossia, 98.93 ± 0.23% [95% CI: 98.51, 99.26] for CLP, and 99.40 ± 0.07% [95% CI: 99.27, 99.52] for the control group. When trained with DP at ε=7.51, which is considered a strong level of privacy in the field (ε<10), the model achieved an average AUROC of 98.73 ± 0.48% [95% CI: 98.59, 98.82] and an accuracy of 95.26 ± 0.90% [95% CI: 94.75, 95.74] across all disorders, conditions, and control groups. Although the differences were statistically significant ($p = 7.56 \times 10^{-10}$), the AUROC values remained close to those of the non-private training, indicating robust performance even under privacy constraints. The reductions in AUROC values ranged from 0.85% to 1.97%, while the decreases in accuracy were less than 5%, demonstrating a very good trade-off between privacy and utility. The DP-trained model maintained high diagnostic performance despite the introduction of privacy-preserving measures. **Supplementary Table 1** provides a comprehensive breakdown of the evaluation results for non-DP and DP training across different ε values, including metrics such as AUROC, accuracy, specificity, and sensitivity for different speech disorders, conditions, and healthy controls.

## 2.2. Guaranteed data privacy compared to conventional model training

To highlight the privacy risks associated with conventional training methods, we conducted a gradient inversion attack following one of the established protocols[17,45]. This attack was applied to both (i) a non-private model and (ii) a private model trained with DP at ε < 10. We used one sample from the dataset to demonstrate the potential risks and protections offered by DP.

**Table 1: Characteristics of the German speech disorder dataset.** The dataset is divided into demographic groups based on sex (female, male) and age: children (0–15 years), young participants (15–30 years), early adults (30–50 years), middle-aged participants (50–70 years), and older participants (70–100 years). Values are provided separately for training and test sets. Each group includes subcategories for healthy controls, Dysarthria, Dysglossia, and Cleft Lip and Palate (CLP) patients. The table reports the total number of speakers and the total recording duration (in hours) for each group. Speech intelligibility is represented by word recognition rates (WRRs), presented as mean ± standard deviation (SD). N/A indicates unavailable data, where some age groups lacked specific speech disorders or conditions.

| Training \| Test | | Dysarthria | Dysglossia | CLP | Control | Overall |
|---|---|---|---|---|---|---|
| **Full dataset** | Speakers [n] | 248 \| 107 | 379 \| 163 | 327 \| 141 | 1025 \| 449 | 1979 \| 860 |
| | Total Duration [h] | 11.83 \| 3.70 | 43.03 \| 20.57 | 26.27 \| 11.83 | 57.61 \| 24.68 | 138.74 \| 60.78 |
| | WRR [mean ± SD] | 69.09 ± 11.44 \| 65.18 ± 16.83 | 64.80 ± 14.35 \| 64.89 ± 12.94 | 46.45 ± 17.24 \| 47.34 ± 17.60 | 64.03 ± 14.05 \| 65.17 ± 13.47 | 61.77 ± 16.09 \| 62.08 ± 15.83 |
| **DEMOGRAPHICS** | | | | | | |
| Female | Speakers [n] | 136 \| 56 | 105 \| 46 | 143 \| 66 | 553 \| 255 | 937 \| 423 |
| | Total Duration [h] | 5.27 \| 1.44 | 20.57 \| 8.96 | 10.01 \| 5.32 | 30.37 \| 13.52 | 66.22 \| 29.24 |
| | WRR [mean ± SD] | 66.20 ± 13.00 \| 73.22 ± 11.33 | 67.10 ± 14.36 \| 73.01 ± 7.90 | 49.41 ± 17.84 \| 45.34 ± 19.06 | 66.50 ± 13.49 \| 68.56 ± 10.77 | 64.26 ± 15.63 \| 67.14 ± 14.53 |
| Male | Speakers [n] | 112 \| 51 | 274 \| 117 | 184 \| 75 | 472 \| 194 | 1042 \| 437 |
| | Total Duration [h] | 6.56 \| 2.26 | 22.46 \| 11.61 | 16.25 \| 6.51 | 27.24 \| 11.16 | 72.51 \| 31.54 |
| | WRR [mean ± SD] | 71.30 ± 9.51 \| 59.80 ± 17.75 | 62.72 ± 14.02 \| 56.28 ± 11.61 | 44.29 ± 16.46 \| 48.90 ± 16.21 | 61.12 ± 14.14 \| 60.86 ± 15.22 | 59.42 ± 16.17 \| 56.78 ± 15.40 |
| [0, 15) years old | Speakers [n] | N/A | N/A | 290 \| 119 | 690 \| 298 | 980 \| 417 |
| | Total Duration [h] | | | 25.08 \| 11.12 | 39.73 \| 17.43 | 64.81 \| 28.55 |
| | WRR [mean ± SD] | | | 45.83 ± 17.28 \| 47.11 ± 16.92 | 63.41 ± 13.74 \| 63.53 ± 13.27 | 57.40 ± 17.20 \| 58.17 ± 16.45 |
| [15, 30) years old | Speakers [n] | N/A | N/A | 37 \| 21 | 325 \| 141 | 362 \| 162 |
| | Total Duration [h] | | | 1.19 \| 0.68 | 17.72 \| 7.08 | 18.91 \| 7.76 |
| | WRR [mean ± SD] | | | 53.73 ± 15.06 \| 47.86 ± 22.23 | 65.21 ± 14.62 \| 68.53 ± 13.24 | 64.88 ± 14.99 \| 66.22 ± 15.89 |
| [30, 50) years old | Speakers [n] | 38 \| 17 | 58 \| 26 | N/A | | 96 \| 43 |
| | Total Duration [h] | 2.57 \| 0.57 | 5.06 \| 0.53 | | | 7.63 \| 1.10 |
| | WRR [mean ± SD] | 68.78 ± 9.42 \| 73.99 ± 14.41 | 60.49 ± 14.50 \| 56.99 ± 14.48 | | | 63.11 ± 13.64 \| 66.86 ± 16.50 |
| [50, 70) years old | Speakers [n] | 95 \| 41 | 237 \| 102 | N/A | | 332 \| 143 |
| | Total Duration [h] | 3.77 \| 1.40 | 26.61 \| 16.66 | | | 30.38 \| 18.06 |
| | WRR [mean ± SD] | 71.55 ± 9.11 \| 58.35 ± 21.68 | 65.12 ± 15.29 \| 65.56 ± 13.44 | | | 65.55 ± 15.01 \| 64.58 ± 14.94 |
| [70, 100) years old | Speakers [n] | 98 \| 43 | 78 \| 32 | N/A | | 176 \| 75 |
| | Total Duration [h] | 5.20 \| 1.56 | 10.86 \| 3.07 | | | 16.06 \| 4.63 |
| | WRR [mean ± SD] | 66.88 ± 13.66 \| 67.82 ± 9.41 | 65.40 ± 10.84 \| 63.03 ± 7.95 | | | 65.72 ± 11.92 \| 64.45 ± 9.32 |

a) ROC curves for different ε values

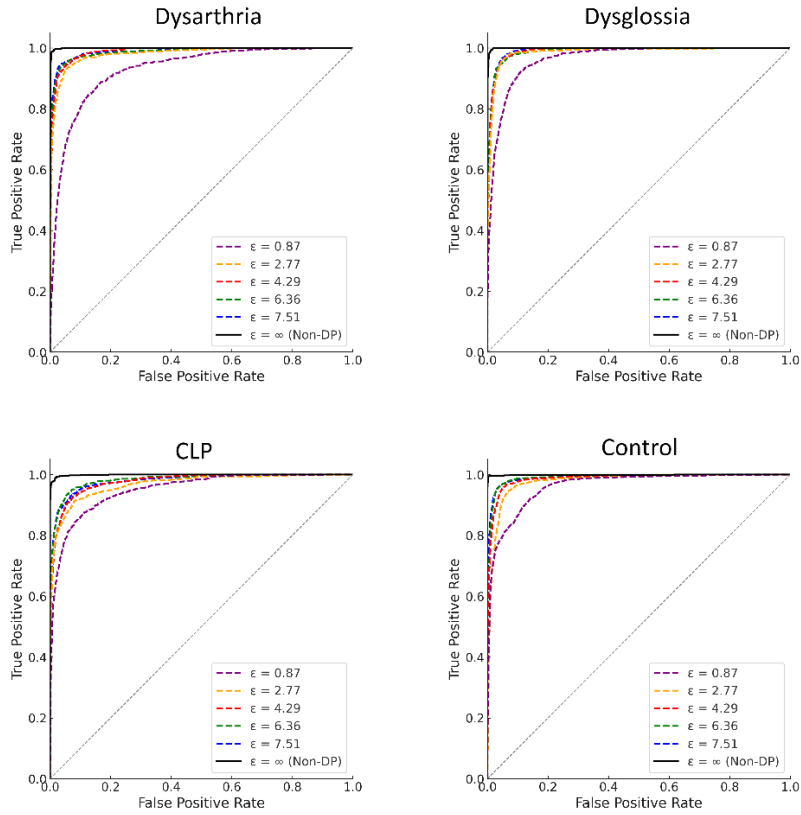

b) Diagnostic accuracy for different ε values

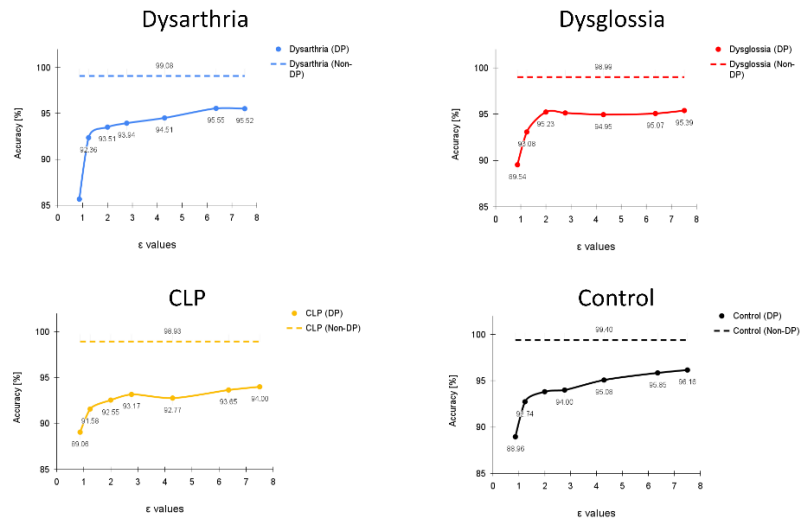

**Figure 2: Diagnostic performance of models trained with and without differential privacy (DP) at different ϵ values for δ = 0.001.** Results are shown for detecting Dysarthria, Dysglossia, Cleft Lip and Palate (CLP), and healthy controls. **(a)** displays the receiver operating characteristic (ROC) curves, where solid lines represent models trained without DP, and dotted lines represent models trained with DP, with different colors corresponding to various ϵ values. The axes depict the true positive rate (sensitivity) versus the False Positive Rate, with the diagonal grey line indicating random chance (no discrimination). **(b)** presents accuracy as a percentage. The training dataset included n=1,979, and the held-out test set comprised n=860 speakers.

**Figure 3** illustrates the results, including the spectrograms and power spectral densities of the original and reconstructed speech signals obtained from leaked network parameters. In the case of the non-private model, the original training speech waveform was reconstructed almost perfectly, notably before the neural network had fully converged. This underscores the serious vulnerabilities of conventional training methods, where sensitive data can be exposed to privacy-focused attacks. Objective quality metrics further support this: the reconstructed signal from the non-private model achieved a signal-to-noise ratio (SNR) of -1.54 dB and a perceptual evaluation of speech quality (PESQ) score of 1.73, indicating partial intelligibility and structure retention.

To demonstrate the efficacy of DP in mitigating such risks, we repeated the experiment with DP applied during training. Using the same gradient inversion approach, no identifiable information could be extracted from the weight updates of the DP-trained model. The reconstructed outputs from the DP-trained model showed no resemblance to the original speech sample, effectively safeguarding patient privacy. This was confirmed quantitatively: the reconstructed signal had a substantially lower SNR of -15.78 dB and a PESQ of 1.15, reflecting high degradation and lack of perceptual similarity to the original audio.

a) Spectrograms

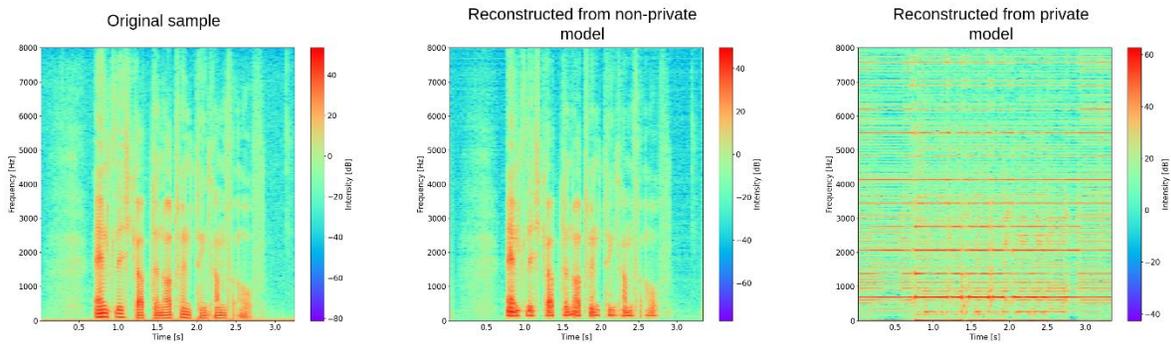

b) Power Spectral Densities

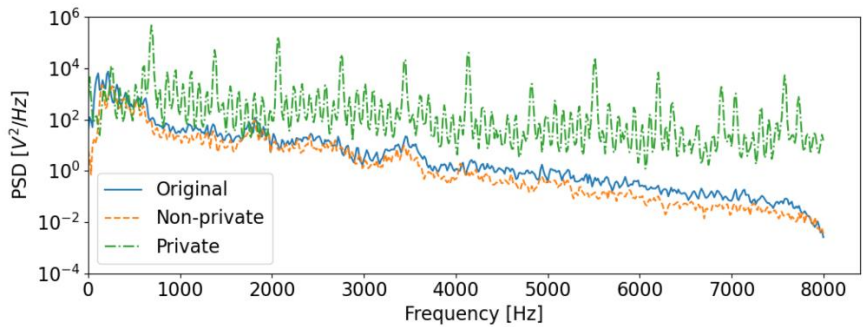

**Figure 3: Spectral representations of a speech sample from the German speech disorder dataset and associated information extraction attacks. (a)** Spectrograms and **(b)** power spectral densities are shown for a 26-year-old male participant from the control group. Results are presented for the original sample, the reconstructed speech from an attacked non-DP model, and a private model with δ = 0.001 and ε < 10. While the non-private model is vulnerable to gradient inversion attacks, allowing reconstruction of the participant's speech from weight updates, the DP-trained model effectively protects sensitive information, resulting in a reconstructed signal that lacks identifiable features.

## 2.3. Under-represented groups are more affected by DP

The German dataset used in this study comprises n=2,983 participants. While this number may seem small compared to image-based datasets, it is important to note that it corresponds to up to 200 hours of recordings, which is considered a very large-scale dataset in the medical speech processing domain[9]. To the best of our knowledge, it is among the largest pathological speech datasets utilized in related publications. Given the known impact of DP on under-represented groups, as reported in the literature, we sought to assess these effects while also exploring whether our findings may extend to a different disorder and language setting. To do so, we used the PC-GITA dataset[44], which consists of speech recordings from a considerably smaller sample of participants (n=50 PD patients and n=50 age- and gender-matched healthy controls), all of whom are native Spanish speakers from Colombia (see **Supplementary Table 2** for dataset characteristics). The task was PD detection.

For the non-private model, the results showed an AUROC of 83.27 ± 1.10% [95% CI: 81.41, 85.15] and an accuracy of 81.75 ± 1.35% [95% CI: 79.52, 84.23]. When trained with DP at $\varepsilon = 7.42$, the AUROC dropped to 73.33 ± 3.87% [95% CI: 67.77, 78.34] and the accuracy to 69.47 ± 3.46% [95% CI: 64.38, 73.75], representing a substantial reduction of up to 12%. This reduction highlights the challenge of maintaining a favorable privacy-utility trade-off, especially for under-represented groups, where the trade-off becomes less effective. To address this issue, and following recent findings in the medical imaging domain[16,24], we applied a slightly more task-specific pretraining. Due to the lack of sufficiently large public datasets for pathological speech, we used a model pre-trained on the train-clean-360 subset (around 360 hours of clean speech) of the LibriSpeech[46] dataset—a widely available healthy speech dataset of English speakers—for weight initialization in the PD detection task. This approach led to a modest performance reduction under privacy constraints compared to the non-private model. For the DP model with $\varepsilon = 4.39$, the AUROC was 80.27 ± 1.06% [95% CI: 78.44, 82.29] and the accuracy was 78.75 ± 1.09% [95% CI: 77.50, 80.00], representing a 3% reduction in both AUROC and accuracy compared to the non-private model. This demonstrates that task-specific pretraining can substantially mitigate the impact of under-representation.

**Figure 4** shows the evaluation results for non-private and DP-trained models across different $\varepsilon$ values, with and without task-specific pretraining, for PD detection. For a comprehensive overview of all evaluation results using all evaluation metrics, refer to **Supplementary Table 3**.

## 2.4. Balancing sex-based fairness under privacy constraints

We evaluated our models based on patient sex and calculated the statistical parity difference (PtD)[47,48] and equal opportunity difference (EOD)[49] to measure fairness. PtD quantifies the difference in diagnostic accuracy between different groups, in this case, between male and female patients. A PtD value of 0 indicates perfect fairness, while positive values suggest a bias favoring one group (e.g., females), and negative values indicate a bias against that group. EOD focuses on fairness in sensitivity, particularly relevant in clinical settings where failing to detect a condition may have serious consequences.

## Diagnostic Performance for PD detection for different ε values

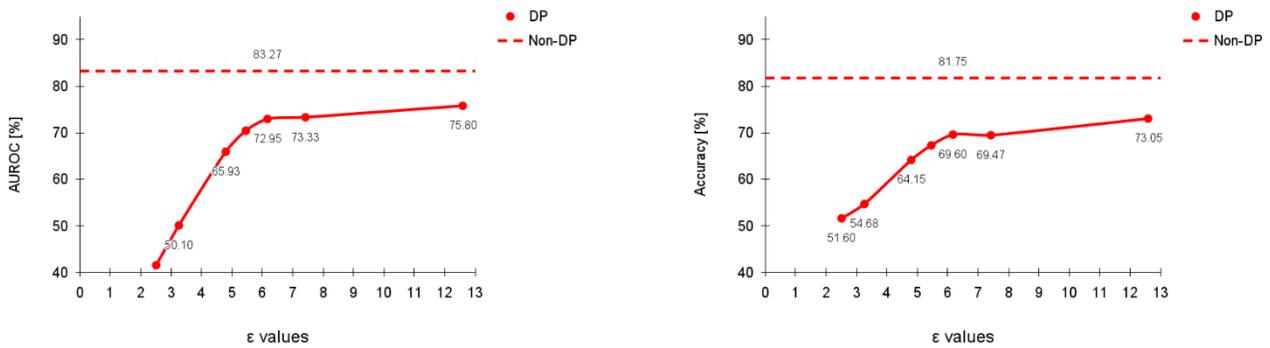

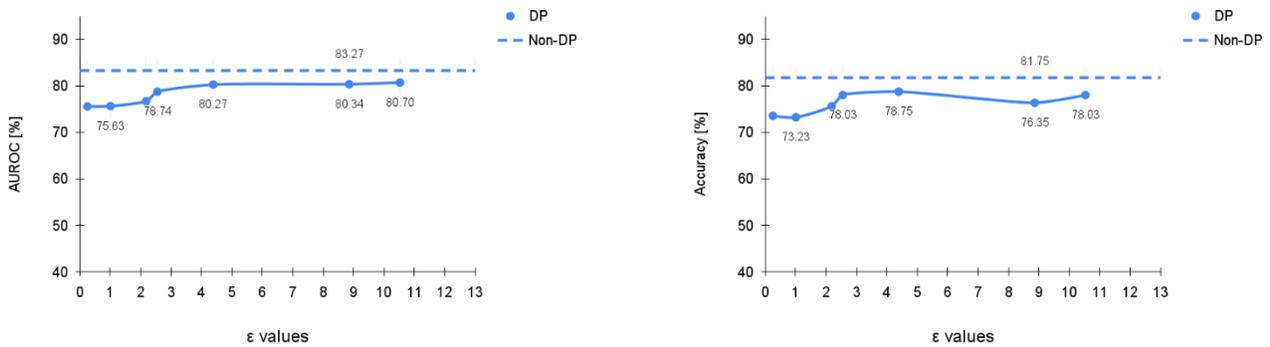

**Figure 4: Evaluation results of differential privacy (DP) training with different ε values for δ = 0.001 on the PC-GITA dataset for Parkinson's Disease (PD) detection, comparing models with and without task-specific pre-trained weights.** The results, shown as percentages, display the area under the receiver operating characteristic curve (AUROC) and accuracy for **(a)** models without task-specific pretraining, using general ImageNet[64] weights, and **(b)** models with task-specific pretraining using weights from the LibriSpeech[46] dataset, a healthy speech dataset of English speakers.

As shown in **Table 2**, diagnostic performance of the non-private model for the female group was slightly higher than for males, with accuracy differences of up to 1.19%, PtD values up to 1.11%, and EOD values up to 1.27% across different speech disorders, conditions, and controls. For the DP model at ε = 7.51, this trend remained consistent, with the model continuing to favor the female groups in all cases, showing PtD values up to 1.87% and EOD values up to 4.65% across the various categories. These results indicated that the privacy-fairness trade-off for sex groups was well-maintained at this privacy level with reasonable privacy-utility trade-off.

However, at extremely high privacy levels, the results differed (see **Figures 5** and **6**). At ε = 0.87, the PtD for Dysarthria increased to 6.51 ± 0.84% (EOD: 4.47 ± 1.00%), showing a disparity. In some groups, such as the control group, the PtD shifted direction, resulting in discrimination against females in favor of males (PtD = -2.72 ± 0.21% and EOD = -9.50 ± 1.00% for females). Additionally, PtD appeared to correlate with privacy levels, as demonstrated by Pearson's correlation coefficients for Dysarthria (r = 0.80), Dysglossia (r = 0.73), CLP (r = 0.55), and control (r = 0.73). These findings suggested that while the privacy-fairness trade-off was well-balanced at reasonable privacy levels (2<ε<10), lower privacy budgets introduced greater discrimination between sex groups, particularly at extremely high privacy levels (ε<2). AUROC values are provided in **Supplementary Table 4**.

**Table 2: Diagnostic accuracy of private and non-private networks across sex groups.** The results, presented as percentages in the format mean ± standard deviation [95% confidence intervals], report the accuracy values for Dysarthria, Dysglossia, Cleft Lip and Palate (CLP), the control group, and the overall average across various ε values with δ = 0.001. These metrics are shown separately for the female (n=423) and male (n=437) subgroups of test speakers. Additionally, statistical parity difference (PtD) and equal opportunity difference (EOD) values for accuracies are included.

|   |   |   | ε = 0.87 | ε = 2.77 | ε = 4.29 | ε = 6.36 | ε = 7.51 | ε = ∞ (Non-DP) |
|---|---|---|---|---|---|---|---|---|
| **Female** | Dysarthria | Accuracy | 89.09 ± 1.14 [86.96, 90.73] | 95.19 ± 0.72 [93.81, 96.42] | 95.53 ± 0.71 [93.92, 96.62] | 96.20 ± 0.59 [94.85, 97.05] | 96.47 ± 0.60 [95.44, 97.62] | 99.58 ± 0.18 [99.18, 99.79] |
|  |  | PtD \| EOD | +6.51 ± 0.84 \| +4.47 ± 1.00 | +2.49 ± 0.23 \| +2.29 ± 0.23 | +1.75 ± 0.25 \| +0.48 ± 0.19 | +1.40 ± 0.09 \| +3.44 ± 0.22 | +1.87 ± 0.09 \| +2.04 ± 0.37 | +1.02 ± 0.06 \| -0.04 ± 0.06 |
|  | Dysglossia | Accuracy | 90.94 ± 1.05 [88.97, 92.67] | 96.08 ± 0.65 [94.63, 97.07] | 95.30 ± 0.89 [93.26, 96.80] | 95.55 ± 0.88 [93.71, 96.89] | 96.12 ± 0.67 [94.91, 97.20] | 99.22 ± 0.32 [98.56, 99.65] |
|  |  | PtD \| EOD | +3.27 ± 0.49 \| +0.10 ± 0.44 | +1.93 ± 0.19 \| +0.77 ± 0.04 | +0.66 ± 0.46 \| +0.38 ± 0.40 | +1.09 ± 0.49 \| +1.33 ± 0.15 | +1.41 ± 0.28 \| +0.43 ± 0.05 | +0.35 ± 0.10 \| +0.25 ± 0.03 |
|  | CLP | Accuracy | 88.94 ± 1.03 [94.96, 95.92] | 94.10 ± 0.83 [92.3, 95.49] | 93.16 ± 0.87 [91.26, 94.52] | 94.42 ± 0.91 [92.12, 95.68] | 94.96 ± 0.64 [93.62, 95.96] | 99.60 ± 0.15 [99.30, 99.82] |
|  |  | PtD \| EOD | -0.83 ± 0.07 \| +3.36 ± 0.48 | +2.07 ± 0.26 \| +2.40 ± 0.36 | +0.27 ± 0.13 \| +1.30 ± 0.29 | +1.23 ± 0.09 \| +2.25 ± 0.06 | +1.51 ± 0.13 \| +4.65 ± 0.34 | +1.11 ± 0.20 \| +1.27 ± 0.38 |
|  | Control | Accuracy | 87.50 ± 0.55 [86.57, 88.52] | 94.36 ± 0.32 [93.77, 94.85] | 95.34 ± 0.26 [94.89, 95.81] | 96.29 ± 0.23 [95.87, 96.66] | 96.48 ± 0.27 [95.95, 96.98] | 99.68 ± 0.08 [99.53, 99.82] |
|  |  | PtD \| EOD | -2.72 ± 0.21 \| -9.50 ± 1.00 | +0.80 ± 0.07 \| +1.35 ± 0.09 | +0.42 ± 0.04 \| +0.47 ± 0.07 | +0.92 ± 0.04 \| +0.85 ± 0.06 | +0.68 ± 0.10 \| +1.16 ± 0.05 | +0.56 ± 0.02 \| +0.94 ± 0.05 |
| **Male** | Dysarthria | Accuracy | 82.58 ± 1.97 [78.78, 86.46] | 92.71 ± 0.94 [91.14, 94.12] | 93.78 ± 0.96 [91.30, 94.82] | 94.80 ± 0.69 [93.51, 95.76] | 94.60 ± 0.69 [93.07, 95.75] | 98.56 ± 0.25 [98.10, 98.94] |
|  |  | PtD \| EOD | -6.51 ± 0.84 \| -4.47 ± 1.00 | -2.49 ± 0.23 \| -2.29 ± 0.23 | -1.75 ± 0.25 \| -0.48 ± 0.19 | -1.40 ± 0.09 \| -3.44 ± 0.22 | -1.87 ± 0.09 \| -2.04 ± 0.37 | -1.02 ± 0.06 \| +0.04 ± 0.06 |
|  | Dysglossia | Accuracy | 87.67 ± 0.57 [86.7, 88.71] | 94.15 ± 0.46 [93.15, 94.82] | 94.64 ± 0.44 [93.73, 95.31] | 94.46 ± 0.40 [93.63, 95.14] | 94.71 ± 0.40 [94.08, 95.50] | 98.88 ± 0.21 [98.39, 99.19] |
|  |  | PtD \| EOD | -3.27 ± 0.49 \| -0.10 ± 0.44 | -1.93 ± 0.19 \| -0.77 ± 0.04 | -0.66 ± 0.46 \| -0.38 ± 0.40 | -1.09 ± 0.49 \| -1.33 ± 0.15 | -1.41 ± 0.28 \| -0.43 ± 0.05 | -0.35 ± 0.10 \| -0.25 ± 0.03 |
|  | CLP | Accuracy | 89.77 ± 0.96 [87.78, 91.33] | 92.03 ± 1.09 [89.87, 94.03] | 92.90 ± 1.00 [90.73, 94.42] | 93.19 ± 0.82 [91.68, 94.42] | 93.45 ± 0.77 [91.75, 94.51] | 98.49 ± 0.34 [97.71, 98.99] |
|  |  | PtD \| EOD | +0.83 ± 0.07 \| -3.36 ± 0.48 | -2.07 ± 0.26 \| -2.40 ± 0.36 | -0.27 ± 0.13 \| -1.30 ± 0.29 | -1.23 ± 0.09 \| -2.25 ± 0.06 | -1.51 ± 0.13 \| -4.65 ± 0.34 | -1.11 ± 0.20 \| -1.27 ± 0.38 |
|  | Control | Accuracy | 90.22 ± 0.35 [89.53, 90.73] | 93.57 ± 0.25 [93.11, 93.93] | 94.93 ± 0.23 [94.50, 95.32] | 95.37 ± 0.20 [95.03, 95.75] | 95.80 ± 0.18 [95.41, 96.14] | 99.12 ± 0.11 [98.92, 99.31] |
|  |  | PtD \| EOD | +2.72 ± 0.21 \| +9.50 ± 1.00 | -0.80 ± 0.07 \| -1.35 ± 0.09 | -0.42 ± 0.04 \| -0.47 ± 0.07 | -0.92 ± 0.04 \| -0.85 ± 0.06 | -0.68 ± 0.10 \| -1.16 ± 0.05 | -0.56 ± 0.02 \| -0.94 ± 0.05 |

# Fairness evaluation based on PtD values

## a) Sex groups

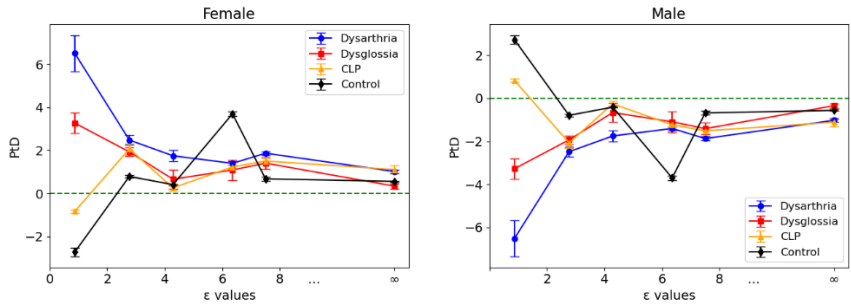

## b) Younger participants

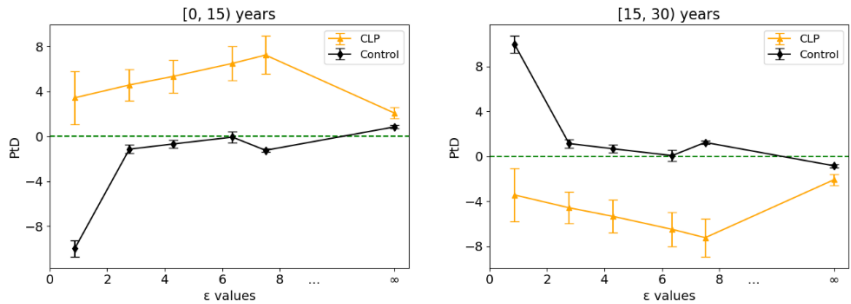

## c) Older participants

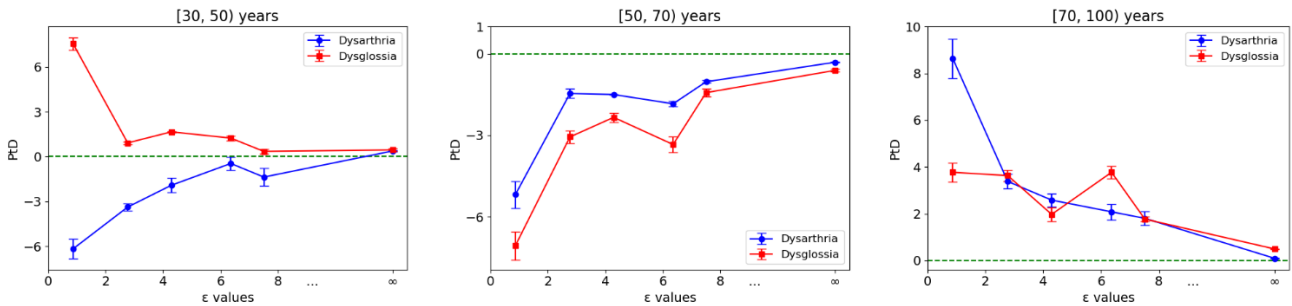

**Figure 5: Statistical parity difference trend at different ε values for δ = 0.001 for demographic fairness evaluation for models trained with and without differential privacy (DP).** The figure shows statistical parity differences (PtD) across demographic subgroups: **(a)** sex groups (female, n=423; male, n=437), and age groups, including **(b)** younger participants ([0, 15) years, n=417; and [15, 30) years, n=162), and **(c)** older participants ([30, 50) years, n=43; [50, 70) years, n=143; and [70, 100) years, n=75). Whiskers represent error bars, showing standard deviation. Dysarthria, Dysglossia, Cleft Lip and Palate (CLP), and control groups are analyzed for sex groups. Due to limited sample sizes, only CLP and control groups are analyzed for younger participants, while only Dysarthria and Dysglossia are analyzed for older participants (see **Table 1** for details).

## 2.5. Age-based privacy-fairness trade-off is more complex under privacy constraints

Next, we evaluated our models based on patient age groups.

# Fairness evaluation based on EOD values

a) Sex groups

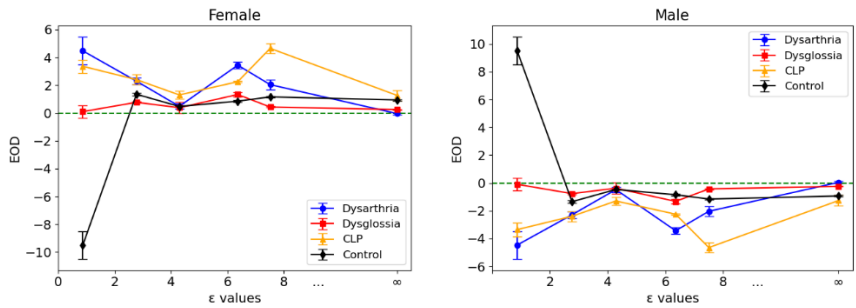

b) Younger participants

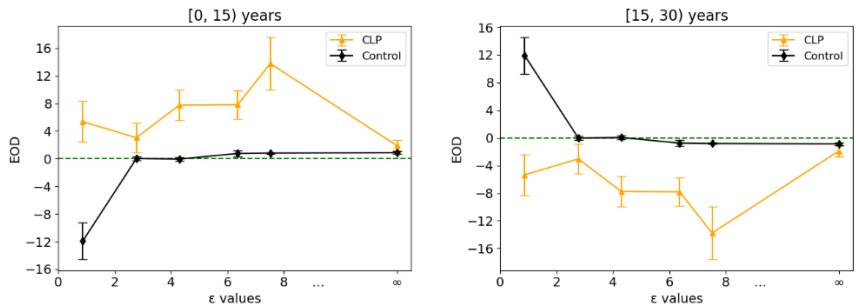

c) Older participants

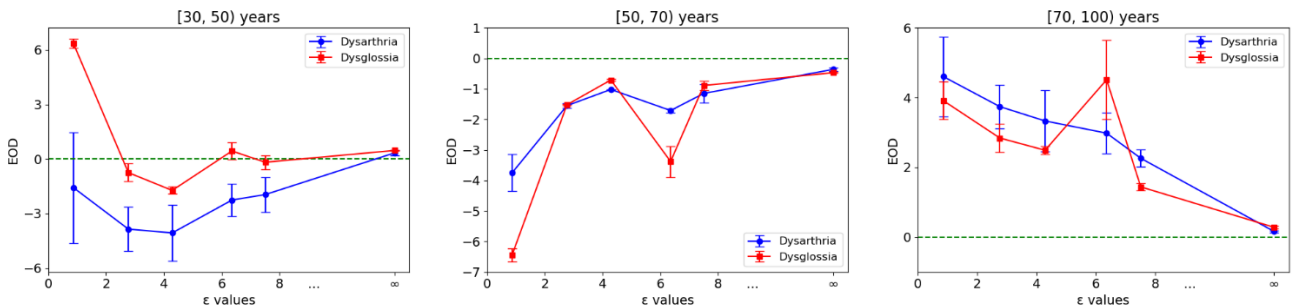

**Figure 6: Equal opportunity difference trend at different ε values for δ = 0.001 for demographic fairness evaluation for models trained with and without differential privacy (DP).** The figure shows equal opportunity differences (EOD) across demographic subgroups: **(a)** sex groups (female, n=423; male, n=437), and age groups, including **(b)** younger participants ([0, 15) years, n=417; and [15, 30) years, n=162), and **(c)** older participants ([30, 50) years, n=43; [50, 70) years, n=143; and [70, 100) years, n=75). Whiskers represent error bars, showing standard deviation. Dysarthria, Dysglossia, Cleft Lip and Palate (CLP), and control groups are analyzed for sex groups. Due to limited sample sizes, only CLP and control groups are analyzed for younger participants, while only Dysarthria and Dysglossia are analyzed for older participants (see **Table 1** for details).

**Table 3** shows the results for age groups. The accuracy of the non-private model was higher for the children (0 to 15 years old) group than for young patients (15 to 30 years old) (99.83% vs. 97.76%), with PtD = +2.07 ± 0.51% and EOD = 1.90 ± 0.76% for CLP. However, for healthy controls, the children group only slightly outperformed the young patients, with a 0.87% difference in accuracy. For detecting healthy controls, a trend similar to that observed with sex groups was maintained. The DP model at ε = 7.51 showed a 4.23% reduction in accuracy and a slight fairness bias in favor of young patients (PtD = 1.23 ± 0.14% and EOD = 0.82 ± 0.08%). At extremely high privacy levels ε < 1, the fairness bias increased, with a similar Pearson's correlation coefficient (r = 0.74).

However, for CLP detection, the results differed (see **Figures 5** and **6**). While the accuracy reduction for the DP model at ε = 7.51 for children was 3.69%, indicating a relatively good trade-off, the reduction for young patients was much larger, at 10.23%. Fairness analysis revealed a major bias towards children compared to young patients, with a PtD of 8.71 ± 2.91% and an EOD of 13.78 ± 3.80%. PtD strongly correlated with privacy levels, as demonstrated by Pearson's correlation coefficient (r = 1.00) for CLP patients between children and young patients.

On the other hand, the reductions in accuracy for Dysarthria and Dysglossia across early adults (30 to 50 years old), middle-aged patients (50 to 70 years old), and older patients (70 to 100 years old) ranged from 4.15% to 6.63% for the DP model at ε = 7.51, indicating a relatively good privacy-utility trade-off for these subgroups. The fairness analysis showed that the non-private model almost did not favor any of these groups over the others for Dysarthria or Dysglossia, with mean PtD values between -1.43% and 1.80% and mean EOD values between -1.96% and 2.26% in all cases. Consistent with the results for sex groups and younger patients, PtD correlated with privacy levels (Pearson's r > 0.75 for all cases except for Dysglossia detection in older patients, where r = 0.55 indicated a moderate correlation). At extremely low privacy budgets, fairness biases were introduced. AUROC values are provided in **Supplementary Table 5**.

## 3. Discussion

In this study, we investigated the impact of differential privacy (DP) on the diagnostic performance of deep learning (DL) models trained on pathological speech data. We focused on the trade-offs between privacy protection and diagnostic accuracy using a large, real-world dataset consisting of approximately 200 hours of recordings from 2,839 German-speaking participants[9,19,42]. This dataset included both pathological and healthy speech samples, covering speech disorders such as Dysarthria and Dysglossia, as well as pathological conditions like Cleft Lip and Palate (CLP).

Our findings demonstrate that private training of diagnostic DL models using pathological speech data is feasible and yields a strong privacy-utility trade-off. With a privacy budget of ε < 10—commonly regarded as a robust level of protection in medical AI[16,24,50,51]—AUROC reductions compared to non-private training ranged from just 0.85% to 1.97%, with accuracy decreases remaining under 5%. These results suggest that DP can preserve patient privacy while maintaining diagnostic utility. Although ε does not correspond to an intuitive unit like accuracy, it quantifies the additional privacy risk from an individual's participation. In practical settings, ε values between 1 and 10—especially when combined with small δ (e.g., $10^{-3}$)—are widely accepted as offering meaningful privacy guarantees[16,24,50,51]. This aligns with regulatory frameworks like the GDPR and the emerging EU AI Act, both of which emphasize demonstrable safeguards against re-identification or "singling out."[16]

**Table 3: Diagnostic accuracy of private and non-private networks across age groups.** The results, presented as percentages in the format mean ± standard deviation [95% confidence intervals], report the accuracy values across various ϵ values with δ = 0.001. Due to an insufficient number of speakers, only Cleft Lip and Palate (CLP) and control groups are analyzed for children and young participants, while only Dysarthria and Dysglossia are analyzed for early adults, middle-aged, and older participants (see **Table 1**). Results are categorized by age groups: children (ages [0, 15), n=417), young participants (ages [15, 30), n=162), early adults (ages [30, 50), n=43), middle-aged (ages [50, 70), n=143), and older participants (ages [70, 100), n=75). Additionally, statistical parity difference (PtD) and equal opportunity difference (EOD) values for accuracies are included.

| | | | ϵ = 0.87 | ϵ = 2.77 | ϵ = 4.29 | ϵ = 6.36 | ϵ = 7.51 | ϵ = ∞ (Non-DP) |
|---|---|---|---|---|---|---|---|---|
| **[0, 15) years old** | CLP | Accuracy | 89.65 ± 0.60 [88.16, 90.40] | 94.01 ± 0.55 [92.91, 95.00] | 94.77 ± 0.50 [93.88, 95.65] | 95.97 ± 0.35 [95.35, 96.73] | 96.14 ± 0.39 [95.25, 96.70] | 99.83 ± 0.07 [99.68, 99.93] |
| | | PtD \| EOD | +3.44 ± 2.39 \| +5.36 ± 2.96 | +4.57 ± 1.44 \| +3.04 ± 2.15 | +5.29 ± 1.38 \| +7.76 ± 2.21 | +7.06 ± 1.75 \| +7.82 ± 2.04 | +8.71 ± 2.91 \| +13.78 ± 3.80 | +2.07 ± 0.51 \| +1.90 ± 0.76 |
| | Control | Accuracy | 85.16 ± 1.76 [82.53, 88.27] | 94.07 ± 0.40 [93.35, 94.86] | 94.81 ± 0.41 [94.12, 95.48] | 95.40 ± 0.38 [94.74, 96.14] | 95.33 ± 0.45 [94.22, 96.09] | 99.56 ± 0.11 [99.38, 99.72] |
| | | PtD \| EOD | -9.97 ± 0.75 \| -11.9 ± 2.66 | -1.15 ± 0.38 \| +0.03 ± 0.29 | -0.68 ± 0.35 \| -0.05 ± 0.29 | -0.07 ± 0.50 \| +0.76 ± 0.46 | -1.23 ± 0.14 \| -0.82 ± 0.08 | +0.83 ± 0.15 \| +0.87 ± 0.19 |
| **[15, 30) years old** | CLP | Accuracy | 86.22 ± 2.97 [80.79, 90.71] | 89.44 ± 1.97 [85.43, 93.24] | 89.48 ± 1.87 [85.27, 92.25] | 88.91 ± 2.09 [83.15, 92.13] | 87.43 ± 3.27 [80.83, 92.32] | 97.76 ± 0.57 [96.52, 98.77] |
| | | PtD \| EOD | -3.44 ± 2.39 \| -5.36 ± 2.96 | -4.57 ± 1.44 \| -3.04 ± 2.15 | -5.29 ± 1.38 \| -7.76 ± 2.21 | -7.06 ± 1.75 \| -7.82 ± 2.04 | -8.71 ± 2.91 \| -13.78 ± 3.80 | -2.07 ± 0.51 \| -1.90 ± 0.76 |
| | Control | Accuracy | 95.13 ± 1.02 [93.24, 96.51] | 95.21 ± 0.78 [93.67, 96.44] | 95.49 ± 0.76 [93.93, 96.76] | 95.47 ± 0.88 [93.84, 96.90] | 96.57 ± 0.58 [95.40, 97.56] | 98.72 ± 0.25 [98.25, 99.19] |
| | | PtD \| EOD | +9.97 ± 0.75 \| +11.9 ± 2.66 | +1.15 ± 0.38 \| -0.03 ± 0.29 | +0.68 ± 0.35 \| +0.05 ± 0.29 | +0.07 ± 0.50 \| -0.76 ± 0.46 | +1.23 ± 0.14 \| +0.82 ± 0.08 | -0.83 ± 0.15 \| -0.87 ± 0.19 |
| **[30, 50) years old** | Dysarthria | Accuracy | 72.70 ± 2.06 [69.59, 76.5] | 89.84 ± 1.06 [87.73, 91.93] | 92.80 ± 1.10 [84.72, 94.85] | 93.82 ± 1.13 [86.03, 96.16] | 92.88 ± 1.25 [90.34, 95.20] | 99.51 ± 0.22 [98.96, 99.74] |
| | | PtD \| EOD | -6.18 ± 0.65 \| -1.58 ± 3.05 | -3.38 ± 0.25 \| -3.86 ± 1.22 | -1.92 ± 0.49 \| -4.07 ± 1.54 | -0.46 ± 0.44 \| -2.26 ± 0.90 | -1.37 ± 0.59 \| -1.96 ± 0.96 | +0.38 ± 0.05 \| +0.34 ± 0.16 |
| | Dysglossia | Accuracy | 83.84 ± 1.03 [82.09, 85.42] | 91.99 ± 0.88 [90.02, 93.49] | 93.05 ± 0.89 [91.46, 94.73] | 92.96 ± 0.71 [91.93, 94.27] | 93.19 ± 0.71 [91.99, 94.53] | 99.11 ± 0.21 [98.7, 99.48] |
| | | PtD \| EOD | +7.56 ± 0.41 \| +6.35 ± 0.26 | +0.92 ± 0.09 \| -0.75 ± 0.49 | +1.66 ± 0.05 \| -1.73 ± 0.20 | +1.24 ± 0.15 \| +0.44 ± 0.47 | +0.35 ± 0.15 \| -0.18 ± 0.39 | +0.45 ± 0.09 \| +0.47 ± 0.02 |
| **[50, 70) years old** | Dysarthria | Accuracy | 74.76 ± 1.81 [71.0, 77.73] | 91.78 ± 0.95 [89.87, 93.62] | 93.51 ± 0.69 [92.38, 94.91] | 93.11 ± 0.82 [91.63, 94.36] | 93.41 ± 0.73 [92.11, 94.53] | 99.01 ± 0.28 [98.41, 99.41] |
| | | PtD \| EOD | -5.18 ± 0.49 \| -3.75 ± 0.61 | -1.46 ± 0.17 \| -1.55 ± 0.08 | -1.50 ± 0.00 \| -1.02 ± 0.00 | -1.84 ± 0.09 \| -1.71 ± 0.07 | -1.03 ± 0.05 \| -1.15 ± 0.31 | -0.31 ± 0.02 \| -0.36 ± 0.06 |
| | Dysglossia | Accuracy | 74.95 ± 1.56 [71.68, 77.23] | 90.11 ± 1.03 [88.23, 92.08] | 90.81 ± 0.91 [89.3, 92.56] | 90.71 ± 0.94 [88.31, 92.23] | 92.38 ± 0.89 [90.15, 93.84] | 98.51 ± 0.30 [97.86, 98.99] |
| | | PtD \| EOD | -7.06 ± 0.52 \| -6.44 ± 0.22 | -3.06 ± 0.22 \| -1.52 ± 0.02 | -2.34 ± 0.17 \| -0.71 ± 0.04 | -3.34 ± 0.28 \| -3.38 ± 0.51 | -1.43 ± 0.15 \| -0.89 ± 0.15 | -0.61 ± 0.05 \| -0.47 ± 0.02 |
| **[70, 100) years old** | Dysarthria | Accuracy | 82.80 ± 1.03 [80.87, 85.02] | 94.59 ± 0.67 [93.28, 95.80] | 95.88 ± 0.53 [95.17, 97.22] | 95.40 ± 0.57 [94.33, 96.50] | 95.05 ± 0.59 [94.00, 96.00] | 99.25 ± 0.27 [98.70, 99.80] |
| | | PtD \| EOD | +8.64 ± 0.85 \| +4.60 ± 1.14 | +3.38 ± 0.31 \| +3.74 ± 0.62 | +2.58 ± 0.28 \| +3.33 ± 0.88 | +2.08 ± 0.34 \| +2.98 ± 0.59 | +1.80 ± 0.29 \| +2.26 ± 0.25 | +0.09 ± 0.01 \| +0.16 ± 0.03 |
| | Dysglossia | Accuracy | 80.53 ± 1.04 [78.7, 82.33] | 94.12 ± 0.76 [93.0, 95.50] | 93.23 ± 0.62 [92.04, 94.42] | 94.94 ± 0.62 [93.74, 96.13] | 94.31 ± 0.76 [92.70, 95.80] | 99.12 ± 0.29 [98.54, 99.67] |
| | | PtD \| EOD | +3.77 ± 0.41 \| +3.91 ± 0.54 | +3.63 ± 0.24 \| +2.84 ± 0.40 | +1.97 ± 0.29 \| +2.49 ± 0.12 | +3.77 ± 0.27 \| +4.51 ± 1.13 | +1.77 ± 0.09 \| +1.44 ± 0.11 | +0.49 ± 0.01 \| +0.28 ± 0.04 |

Building on evidence from the medical imaging domain[16,21,24,51], which suggests that DP performs best with large datasets, we tested its effectiveness on a smaller dataset. To preliminarily evaluate the robustness of our findings, we applied our approach to the PC-GITA dataset[44], consisting of n=100 Spanish-speaking participants, for the task of PD detection. This dataset allowed us to evaluate DP's impact in a different language and for a neurological disorder rather than a speech disorder. As expected, the smaller dataset led to a substantial 12% reduction in performance, undermining the privacy-utility trade-off with DP training. However, by pretraining on a large-scale public healthy speech dataset, we were able to mitigate these performance losses to a substantial degree. Despite this improvement, we recognize that access to large-scale medical speech datasets remains more challenging compared to medical imaging, where multiple public datasets, such as MIMIC-CXR[52] and CheXpert[53], are readily available. We anticipate that even smaller performance reductions could be achieved if pretraining was performed on a large-scale pathological speech dataset specifically. While our results indicate that task-specific pretraining on LibriSpeech[46] can mitigate performance degradation under DP, we acknowledge a potential domain mismatch due to differences in language (English vs. Spanish), speaker demographics, and the presence of pathology. These factors may limit the generalizability of learned representations. Nevertheless, our findings suggest that even general-purpose healthy speech can serve as a useful pretraining source when task-specific pathological data are limited. We encourage future studies to explore more linguistically and clinically aligned pretraining corpora, once available, to further improve transferability and reduce bias in multilingual pathological speech applications. We advocate for further research into data-sharing approaches, such as automatic speaker anonymization[19], to facilitate the public release of large-scale datasets in the healthcare speech domain. This would further advance DP development and eventually improve patient outcomes[54].

Given the concerns raised in the literature about DP's differential impact on demographic subgroups[16,24,29,32], we conducted a detailed analysis of both sex and age groups. The results were intriguing. For sex groups, DP had a minimal effect on the privacy-fairness trade-off at commonly accepted privacy levels, with accuracy differences of up to 1.19%. The original bias, where females were generally easier to diagnose than males in the non-private model, remained consistent under DP. This trend was also observed across most age groups, where DP did not exacerbate or reduce the original biases of the non-private model, except for CLP detection in young patients (15 to 30 years old). Further investigation revealed that this group had a small sample size (n=21), making the results less reliable. Future studies should explore the performance of private models for CLP detection among young patients using larger, more representative datasets.

In contrast to the image domain[16,21,24,51], where diagnostic accuracy among demographic subgroups showed little correlation with privacy levels, our study revealed a relatively strong correlation (Pearson's r > 0.70 in most cases). This indicates that as privacy levels become more stringent ($\epsilon < 1$), significant disparities emerge. For example, in Dysarthria detection, we observed a PtD of 6.51 ± 0.84% for sex groups and 8.64 ± 0.85% for age groups at $\epsilon = 0.87$. While pretraining is crucial, these findings highlight the importance of selecting an appropriate privacy budget. The privacy-fairness trade-off is not linear and pushing for extremely low privacy budgets ($\epsilon < 1$) can lead to substantial discrimination among subgroups. However, the good news is that for privacy levels within the range of $1 < \epsilon < 10$, which are generally considered safe[16,50], the trade-off remains almost linear and consistent with results from other domains.

To highlight the tangible privacy risks of conventional model training, we conducted a proof-of-concept gradient inversion attack. While our demonstration was limited to a single representative sample due to the substantial computational cost of gradient-based speech reconstruction—particularly the vocoding step—we were able to reconstruct identifiable speech with high fidelity from a non-private model early in training. This example was chosen to illustrate that speech models trained without DP remain vulnerable to information leakage. Although not designed to report aggregate attack success rates, this real-world reconstruction underscores the importance of integrating formal privacy mechanisms like DP, which, in our evaluation, suppressed the ability to extract any intelligible speech from the model's gradients. Future work could systematically benchmark attack success rates across larger subsets to quantify the generalizability of these risks.

Our study has several limitations. First, the speech disorder dataset we utilized has some constraints. Specifically, not all age groups had sufficient speech samples or participants across the various speech disorders and pathological conditions analyzed. For instance, subgroups such as participants aged 15–30 years were relatively small, which may limit the statistical reliability of fairness assessments across age groups. However, to the best of our knowledge, this dataset is the largest pathological speech dataset used in published studies, encompassing recordings collected over an extended period from multiple institutions. Due to privacy regulations, the dataset is not yet publicly available, though a representative sample is provided (see **Supplementary Data 1**). This limitation restricted our ability to thoroughly analyze all subgroups across all disorders. Future research should aim to address these gaps by utilizing larger and more diverse datasets. Second, the generalizability of our results has certain limitations: (i) We only considered German and Spanish language datasets. While the task of speech disorder detection is generally considered speaker-level and language-independent, future studies should validate our findings using datasets from additional languages to ensure broader applicability. (ii) We focused exclusively on CNN-based[55] models for DP training. This decision was guided by extensive prior literature[16,24,50,51] demonstrating successful and stable application of DP-SGD with CNN architectures in various medical imaging tasks. Furthermore, recent work[56] comparing CNNs and vision-transformer-based models in a DP setting showed that transformer-based models exhibit substantially higher utility degradation. Given the lack of maturity in DP methods for transformers, we prioritized CNNs to ensure methodological stability and comparability. Nevertheless, we advocate for future research exploring DP-compatible transformer models tailored to speech-based clinical tasks. Third, while we used speech disorder detection as our primary measure of utility, we recognize that this approach may only scratch the surface of understanding DP's broader impacts on pathological speech data. Future studies should explore more complex clinical tasks involving pathological speech, utilizing DP training to assess its effects in greater detail. Lastly, we acknowledge that achieving training convergence in DP AI models is a more challenging and computationally intensive process[16,21,24,43]. To support the research community, we are publicly providing access to our comprehensive framework and source code, along with recommended configurations for more efficient DP training.

In conclusion, our study demonstrates the feasibility of applying DP to AI models in the context of pathological speech analysis. We achieved a very good balance between privacy protection and diagnostic performance, showing that DP can protect sensitive patient data while maintaining high accuracy, particularly with large datasets. The application of task-specific pretraining proved essential for mitigating performance losses in smaller datasets, underscoring the importance of tailored approaches in privacy-preserving AI. As the field of AI-driven healthcare continues to evolve, integrating DP into diverse clinical applications will be crucial for ensuring both patient privacy and equitable outcomes. Our work paves the way for future research in this area, aiming to refine DP methodologies and extend their impact across various medical domains, ultimately contributing to the development of safer and more trustworthy AI technologies in healthcare.

# 4. Methods

## 4.1. Ethics statement

The German dataset received approval from the institutional review board of University Hospital Erlangen under application number 3473, in compliance with the Declaration of Helsinki. The protocol for the PC-GITA dataset[44] was approved by the Ethical Committee of the Research Institute in the Faculty of Medicine at the University of Antioquia in Medellín, Colombia (approval number 19-63-673). All experiments were conducted in accordance with applicable national and international guidelines and regulations and informed consent was obtained from all adult participants, as well as from the parents or legal guardians of the children involved for both datasets.

## 4.2. Datasets

### 4.2.1. Speech disorders dataset

The speech disorders dataset[9,19,42] used in this study encompasses a broad spectrum of speech samples collected from various locations throughout Germany. Participants had a mean age of 30 ± 25 [standard deviation] years, with a balanced representation of both male and female participants and included individuals ranging from children to elderly adults. **Table 1** details the demographic characteristics of the dataset. The dataset includes participants with Dysarthria and Dysglossia, as well as corresponding healthy controls. These participants were tasked with reading "Der Nordwind und die Sonne,"[42] a phonetically diverse German adaptation of Aesop's fable "The North Wind and the Sun," which consists of 108 words, 71 of which are unique. For participants with CLP and their healthy controls, the "Psycholinguistische Analyse kindlicher Sprechstörungen" (PLAKSS)[57] test was carried out, requiring them to name pictograms presented on slides, covering all German phonemes in various positions. To handle the tendency of some children to use multiple words or add extra words between target phrases, recordings were automatically segmented at pauses longer than one second[19,42].

Data collection spanned from 2006 to 2019, primarily during routine outpatient examinations at the University Hospital Erlangen, as well as from over 20 other locations across Germany for control speakers, resulting in a diverse range of regional dialects among participants. All participants were informed about the study's objectives and procedures, and consent was obtained. A standardized recording protocol ensured consistent microphone setups and speech tasks across all sessions. The study excluded non-native speakers and individuals whose speech was significantly impacted by factors unrelated to the targeted disorders. The dataset was managed using the PEAKS[42] software, a widely recognized open-source tool in the German-speaking research community and recordings were made at a 16 kHz sampling rate and 16-bit resolution[9,19]. For this study, we followed the exclusion criteria shown in **Figure 7**, ensuring speech quality and noise standards, and removing recordings with multiple speakers.

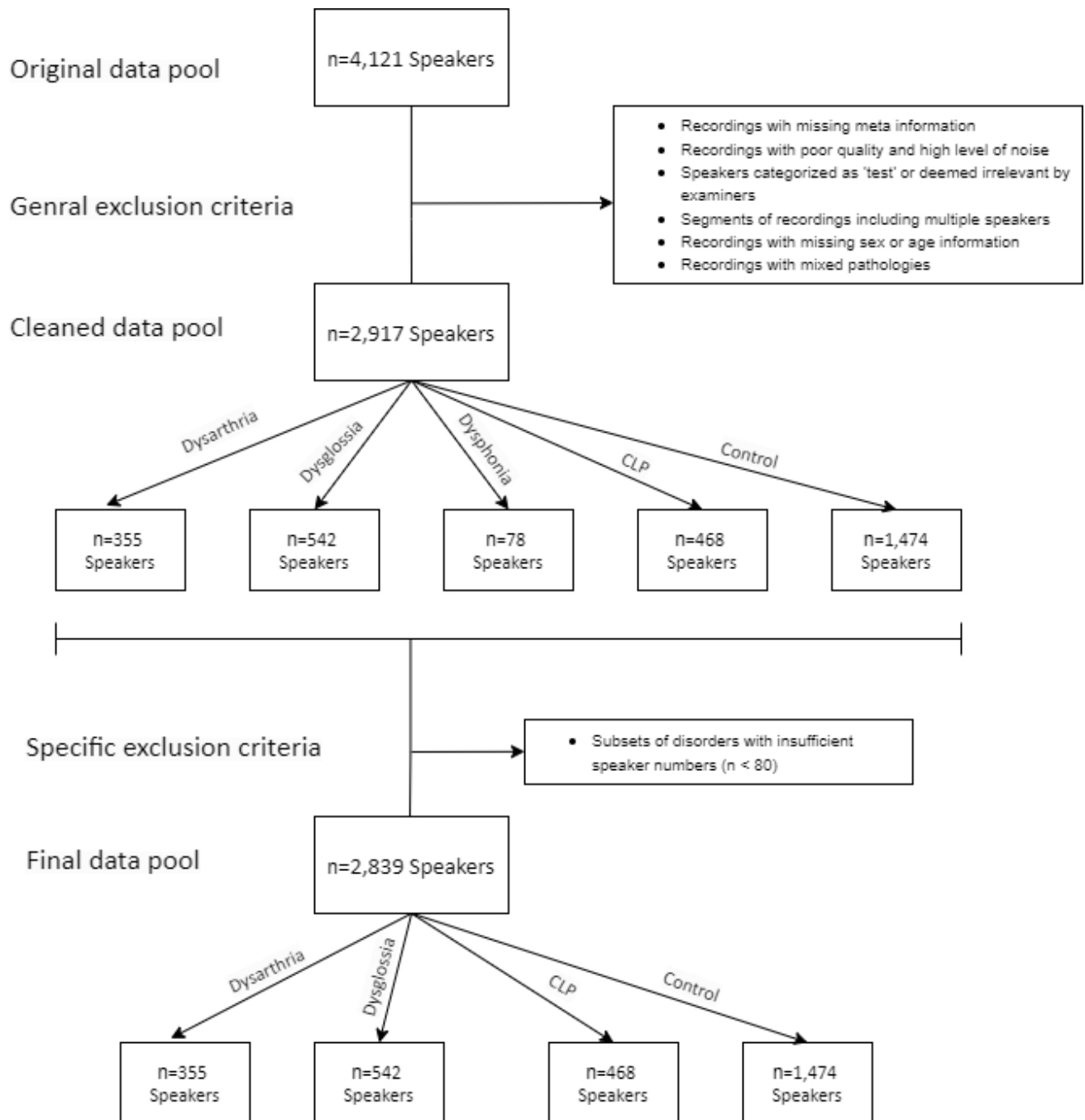

Figure 7: Flowchart of the inclusion and exclusion criteria for the speech disorder dataset for this study.

### *4.2.2. Parkinson's Disease dataset*

The PC-GITA dataset[44] was used for Parkinson's disease (PD) detection, comprising speech recordings from 50 PD patients and 50 healthy controls, all of whom are native Colombian Spanish speakers. The male PD patients ranged in age from 33 to 77 years (mean 62.2 ± 11.2 years), while female PD patients ranged from 44 to 75 years (mean 60.1 ± 7.8 years). Among the healthy controls, the men were aged 31 to 86 years (mean 61.2 ± 11.3 years), and the women were aged 43 to 76 years (mean 60.7 ± 7.7 years)[44]. This ensures that the dataset is well balanced in terms of age and gender. All recordings were made in controlled noise conditions within a soundproof booth, and the audio was sampled at 44.1 kHz with 16-bit resolution. None of the healthy controls exhibited symptoms of PD or any other neurological disorders. **Supplementary Table 2** details the demographic characteristics of the dataset. Additional details about the dataset can be found in the original publication[44].

### *4.2.3. LibriSpeech dataset*

The LibriSpeech[46] dataset is a large-scale public corpus of English speech derived from audiobooks in the LibriVox project. It contains approximately 1,000 hours of speech, organized into training, development, and test sets. The data is further categorized into "clean" (recordings with minimal background noise and speaker variation) and "other" (which includes more challenging conditions) subsets. All recordings are sampled at 16 kHz. For this study, we used the largest subset, train-clean-360, comprising 363 hours of clean speech from 921 speakers (n=439 females and n=482 males), for the general task of detecting gender, as the purpose was pretraining for a diagnostic model.

## 4.3. Differential privacy for deep learning

Differential Privacy (DP) is a formal framework that quantifies the privacy guarantees of algorithms trained on sensitive data[21,22,50]. A randomized algorithm $M: X \to y$ satisfies (ε, δ)-DP if, for any two datasets $D_1$ and $D_2$ differing in a single record and for any subset of outputs $S \subseteq Range(M)$, the following holds:

$$P(M(D_1) \in S) \leq e^\varepsilon . P(M(D_2) \in S) + \delta \qquad (1)$$

Here, ε measures the privacy loss (smaller values imply stronger privacy), while δ represents the probability that this guarantee may not hold in rare cases.

In deep learning, DP is commonly implemented using the Differentially Private Stochastic Gradient Descent (DP-SGD) algorithm[43]. DP-SGD modifies standard SGD by (i) clipping per-sample gradients to a fixed norm, limiting the influence of any single data point, and (ii) adding Gaussian noise to the aggregated gradients based on a calibrated noise scale that corresponds to the desired (ε, δ) budget. This ensures that updates during training do not reveal sensitive individual-level information[24].

Privacy accounting techniques such as the moments accountant are used to track cumulative privacy loss across training steps and ensure the final model remains within the defined privacy budget[43].

## 4.4. Experimental design

Two distinct networks, specifically, models trained employing DP- or non-DP training, were trained on the same training dataset. Subsequently, testing was performed on a separate held-out test set for both networks, resulting in a strictly paired comparison scenario, thereby removing the need for random effects modeling. It should be noted that we used a multiclass classification approach, optimizing for average performance across all classes and did not perform a detailed comparison for individual disorders. In this study, we considered per-patient privacy.

## 4.5. Deep learning network architecture and training

### 4.5.1. Data preprocessing

During the data preprocessing phase, any drifting noise present in the audio was removed using a forward-backward filter[58]. The final feature set consisted of 80-dimensional log-Mel-spectrograms, generated using a short-time Fourier transform with a size of 1024[19], a window length of 64 ms, 80 Mel filters, and a frequency range of 0 to 8 kHz.

### 4.5.2. Network architecture

Given the two-dimensional structure of log-Mel-spectrograms (see **Supplementary Figure 1** for examples representing a speech sample from a different speaker for each condition: Dysarthria, Dysglossia, CLP, and healthy participants) and the compatibility of the original DP-SGD algorithm with CNNs rather than transformers for training, we selected the ResNet18[55] model, originally designed for image classification, to enhance feature extraction[59,60]. Notably, previous research[16,24,51], particularly in the medical imaging domain, has demonstrated the generalizability of the DP-SGD algorithm across various CNNs for medical diagnostics. To ensure compatibility with DP-SGD training, we used a modified ResNet18[55] architecture incorporating adjustments proposed by Klause et al.[61]. Instead of batch normalization[62], we employed group normalization[63] with groups of 32, which is more appropriate for DP settings. The network's inputs were 3-channel Mel-filterbank energies, aligned with the pretrained weights from the large-scale ImageNet[64] dataset, consisting of over 14 million images across 1,000 categories. The first layer of the network outputted 64 channels, and the final fully connected layer reduced the 512 extracted features to match the number of targeted disorders. To convert output predictions to class probabilities, we applied the logistic sigmoid function.

### *4.5.3. Non-DP training*

For non-DP training, we used a batch size of 128, with 8 utterances per speaker randomly selected for each batch. For the experiments with the smaller PC-GITA dataset, we adjusted the batch size to 40, with utterances per speaker selected accordingly. To accommodate the varying lengths of log-Mel-filterbank energies, we randomly selected 180 frames for inclusion in the training process. The network inputs were structured as at $128 \times 3 \times 80 \times 180$ for the speech disorder dataset and $40 \times 3 \times 80 \times 180$ for the PC-GITA dataset, corresponding to batch size, channel size (adjusted to 3 to match the expectations of the pretrained network by replicating log-Mel-spectrograms three times), log-Mel-spectrogram dimensions, and frame size. The training process was conducted over 200 epochs, utilizing the Adam[65] optimizer with a learning rate of $5 \times 10^{-5}$. Binary weighted cross-entropy was used for loss calculation[19].

### *4.5.4. DP training*

For DP training, all models were optimized using the NAdam optimizer[66] with a learning rate of $5 \times 10^{-4}$ to achieve optimal convergence, without applying weight decay. Binary cross-entropy was selected as the loss function. The maximum allowed gradient norm was set to 1.5, which was determined to be optimal for this context. Each data point in the DP training batches was sampled with a probability equal to the batch size (128 for the speech disorder dataset and 40 for the PC-GITA dataset) divided by the total number of training samples in the dataset. A DP accountant, based on Rényi differential privacy[67], was employed to manage the privacy budget (represented by ε and δ) and ensure it remained within predetermined limits. A δ value of 0.001 was chosen for all networks. The value of ε depended on factors such as the introduced noise, the set δ, the number of training steps, and the batch size. The reported ε was determined by the convergence step of each neural network, given the diversity of the datasets[24].

## 4.6. Information attack method

The information attack was conducted following the same preprocessing steps described earlier, converting speech waveforms into log-Mel-spectrograms for training in a diagnostic convolutional neural network. The improved Deep Leakage from Gradients (iDLG)[45] method was employed as the attack strategy. For proof-of-concept purposes, a lightweight LeNet[68] model was used as the primary speech diagnosis network. Two paired training scenarios were designed: (i) without applying privacy measures and (ii) with DP training at δ = 0.001 and ε < 10.

During the training of the diagnostic model using pathological speech data, gradients were extracted in the early stages of training to simulate information leakage. These gradients served as inputs to a generative unsupervised model, as proposed in[45]. Dummy data, initialized with random values matching the dimensions of the input data, were iteratively optimized using the Limited-memory Broyden–Fletcher–Goldfarb–Shanno (L-BFGS) algorithm[69], a cross-entropy loss function, and a learning rate of 0.1 to reconstruct the original training data from the leaked gradients. Our experiments demonstrated that 100 iterations were sufficient for convergence, yielding log-Mel-spectrograms as the reconstructed data. To synthesize speech waveforms from the reconstructed Mel-spectrograms, the HiFi-GAN[70] vocoder was employed. This state-of-the-art voice synthesizer was pre-trained on the LJ Speech[71] corpus. This publicly available dataset comprises 13,100 short audio clips from a single speaker reading passages from seven non-fiction books.

## 4.7. Evaluation

### *4.7.1. Privacy-utility trade-off*

Accuracy and the area under the receiver operating characteristic curve (AUROC) were the primary evaluation metrics used to assess diagnostic performance, with outcomes for individual disorders averaged without applying weights. To analyze the privacy-utility trade-off, ε was used as the privacy measure, while both accuracy and AUROC served as utility measures. Sensitivity and specificity were calculated as secondary metrics for diagnostic performance. For sensitivity and specificity calculations, the threshold was determined using Youden's criterion[72], which maximizes the difference between the true positive rate and the false positive rate.

For the speech disorder dataset, speakers were randomly allocated to training (70%) and test (30%) groups. This random allocation was consistent across experiments to ensure that the same training and test subsets were used when comparing, allowing for paired analyses that account for random variations. The division was designed to prevent overlap between training and test data. The final training set included n=1,979 speakers, and the final test set included n=860 speakers. A similar procedure was followed for the PC-GITA dataset, resulting in a final training set of n=80 speakers and a test set of n=20 speakers.

### *4.7.2. Privacy-fairness trade-off*

To evaluate the privacy-fairness trade-off, we assessed the performance of private and non-private networks across different demographic subgroups. Detailed demographic information is provided in **Table 1**.
Additionally, we calculated two fairness metrics: statistical parity difference (PtD)[47,48] and equal opportunity difference (EOD)[49].
PtD is defined as:

$$PtD = P(\hat{Y} = 1 | C = Minority) - P(\hat{Y} = 1 | C = Majority) \qquad (2)$$

where $\hat{Y} = 1$ represents the model's correct predictions, and $C$ is the demographic group in question[16]. PtD quantifies the difference in diagnostic accuracy between groups, such as male and female patients. A PtD value of 0 indicates perfect fairness, while positive values suggest a bias favoring one group, and negative values indicate a bias against that group.

EOD was calculated as the difference in true positive rates (TPR) between demographic groups:

$$EOD = TPR_{Minority} - TPR_{Majority} \qquad (3)$$

where TPR is defined as the proportion of correctly identified positive cases out of all actual positive cases for each group. EOD focuses on fairness in sensitivity, particularly relevant in clinical settings where failing to detect a condition may have serious consequences.

To examine the relationship between fairness results and privacy levels, Pearson's correlation coefficient was employed.

The demographic subgroups considered in this study included sex groups (females vs. males) and age groups, categorized as children (0 to 15 years old), young patients (15 to 30 years old), early adults (30 to 50 years old), middle-aged patients (50 to 70 years old), and older patients (70 to 100 years old).

### *4.7.3. Statistical analysis*

Statistical analyses were conducted using Python (v3) along with the SciPy and NumPy packages. Given that each speaker provided multiple utterances and to address the randomness in sampling during training and testing, each test phase was repeated n=50 times to minimize potential biases. Evaluations were carefully paired to ensure consistent comparisons between DP and non-DP scenarios. The results are presented as mean ± standard deviation [95% confidence intervals]. Statistical significance was assessed using a two-tailed Wilcoxon signed-rank test[73]. The family-wise alpha threshold was set at 0.05[24].

## 4.8. Data availability

The German speech disorder dataset used in this study is internal data of patients of the University Hospital Erlangen and is not publicly available due to patient privacy regulations. A reasonable request to the corresponding author is required for accessing the data on-site at the University Hospital Erlangen in Erlangen, Germany. The PC-GITA[44] dataset is a restricted-access resource. To gain access, users must agree to the dataset's data protection requirements by submitting a request to JROA (rafael.orozco@udea.edu.co). The LibriSpeech[46] dataset is publicly available at https://www.openslr.org/12 under a CC BY 4.0 license. The LJ Speech[71] dataset is publicly available at https://keithito.com/LJ-Speech-Dataset/.

## 4.9. Code availability

To encourage transparency and facilitate future research, we have publicly released our complete source code at https://github.com/tayebiarasteh/DPSpeech. The repository provides comprehensive documentation on the training procedures, evaluation protocols, and data preprocessing steps used in our study. This will enable the research community to reproduce our results effectively. The code is implemented in Python 3.9 and leverages the PyTorch 1.13 framework for all deep learning operations.

# Additional information

## Acknowledgements

We acknowledge financial support by Deutsche Forschungsgemeinschaft (DFG) and Friedrich-Alexander-Universität Erlangen-Nürnberg within the funding programme "Open Access Publication Funding."

## Author contributions

The formal analysis was conducted by STA, AM, and SHY. The original draft was written by STA and corrected by STA, PAPT, AM, and SHY. The software was developed by STA. The experiments were performed by STA, ML, and MR. Statistical analysis was performed by ML and STA. Datasets were provided by JROA, MS, AM, and SHY. STA cleaned, organized, and pre-processed the German data. JROA cleaned, organized, and pre-processed the Spanish data. STA and MS provided clinical expertise. STA, ML, PAPT, TAV, MR, JROA, AM, and SHY provided technical expertise. STA designed the study. All authors read the manuscript, contributed to the editing, and agreed to the submission of this paper.

# Supplementary Information

**Supplementary Table 1: Detailed evaluation results for training with and without differential privacy (DP).** The table presents the area under the receiver operating characteristic curve (AUROC), accuracy, specificity, and sensitivity, expressed as percentages in the format mean ± standard deviation [95% confidence intervals], for each disorder and the control group from the German speech disorder dataset (refer to **Table 1** for more details) for different ε values and δ = 0.001. The training dataset comprised n=1,979 speakers, while the held-out test set included n=860 speakers.

| ε | Disorder | AUROC | Accuracy | Specificity | Sensitivity |
|---|---|---|---|---|---|
| ∞ (Non-DP) | Dysarthria | 99.90 ± 0.02 [99.86, 99.93] | 99.08 ± 0.18 [98.75, 99.42] | 99.07 ± 0.22 [98.67, 99.47] | 99.18 ± 0.28 [98.6, 99.62] |
| | Dysglossia | 99.94 ± 0.01 [99.93, 99.96] | 98.99 ± 0.14 [98.69, 99.19] | 98.95 ± 0.17 [98.58, 99.23] | 99.18 ± 0.22 [98.79, 99.62] |
| | CLP | 99.91 ± 0.03 [99.86, 99.95] | 98.93 ± 0.23 [98.51, 99.26] | 98.96 ± 0.31 [98.36, 99.39] | 98.78 ± 0.30 [98.2, 99.41] |
| | Control | 99.91 ± 0.01 [99.89, 99.93] | 99.40 ± 0.07 [99.27, 99.52] | 99.54 ± 0.13 [99.31, 99.75] | 99.27 ± 0.12 [99.03, 99.44] |
| 0.87 | Dysarthria | 92.81 ± 0.37 [92.11, 93.54] | 85.67 ± 0.95 [83.94, 86.89] | 85.44 ± 1.24 [83.2, 87.09] | 87.27 ± 1.46 [84.63, 90.37] |
| | Dysglossia | 96.46 ± 0.14 [96.25, 96.75] | 89.54 ± 0.60 [88.6, 90.59] | 88.79 ± 0.93 [87.29, 90.50] | 92.74 ± 0.95 [91.06, 94.53] |
| | CLP | 94.94 ± 0.17 [94.55, 95.19] | 89.06 ± 0.65 [87.89, 90.37] | 89.47 ± 0.97 [87.81, 91.39] | 86.97 ± 1.17 [84.71, 88.81] |
| | Control | 96.14 ± 0.10 [95.92, 96.29] | 88.96 ± 0.21 [88.57, 89.34] | 87.44 ± 1.43 [85.14, 90.49] | 90.35 ± 1.49 [87.27, 92.78] |
| 2.77 | Dysarthria | 98.06 ± 0.17 [97.78, 98.43] | 93.94 ± 0.81 [91.99, 95.31] | 93.92 ± 1.05 [91.56, 95.80] | 94.08 ± 1.11 [91.97, 95.53] |
| | Dysglossia | 98.79 ± 0.11 [98.59, 99.0] | 95.13 ± 0.44 [94.43, 95.91] | 94.82 ± 0.69 [93.64, 96.07] | 96.42 ± 0.75 [94.97, 97.91] |
| | CLP | 96.91 ± 0.19 [96.59, 97.31] | 93.17 ± 0.60 [92.13, 94.30] | 93.97 ± 0.91 [92.45, 95.64] | 89.07 ± 1.24 [86.54, 91.09] |
| | Control | 97.70 ± 0.12 [97.50, 97.89] | 94.00 ± 0.18 [93.65, 94.35] | 92.94 ± 0.63 [92.01, 94.26] | 94.98 ± 0.61 [93.8, 95.90] |
| 4.29 | Dysarthria | 98.57 ± 0.12 [98.37, 98.81] | 94.51 ± 0.74 [93.05, 95.67] | 94.55 ± 0.91 [92.83, 96.06] | 94.23 ± 0.87 [92.64, 95.88] |
| | Dysglossia | 98.94 ± 0.07 [98.81, 99.05] | 94.95 ± 0.40 [94.24, 95.68] | 94.59 ± 0.61 [93.57, 95.81] | 96.49 ± 0.71 [95.05, 97.61] |
| | CLP | 97.52 ± 0.13 [97.27, 97.76] | 92.77 ± 0.77 [91.47, 94.31] | 93.08 ± 1.13 [91.22, 95.3] | 91.22 ± 1.21 [88.82, 93.13] |
| | Control | 98.53 ± 0.07 [98.41, 98.65] | 95.08 ± 0.16 [94.8, 95.37] | 94.71 ± 0.51 [93.71, 95.67] | 95.42 ± 0.49 [94.24, 96.16] |
| 6.36 | Dysarthria | 98.77 ± 0.11 [98.56, 98.94] | 95.55 ± 0.59 [94.08, 96.33] | 95.72 ± 0.75 [93.83, 96.74] | 94.33 ± 0.81 [92.72, 95.77] |
| | Dysglossia | 98.95 ± 0.07 [98.78, 99.05] | 95.07 ± 0.44 [94.16, 95.90] | 94.79 ± 0.68 [93.53, 95.98] | 96.29 ± 0.77 [95.11, 97.45] |
| | CLP | 98.15 ± 0.11 [97.96, 98.28] | 93.65 ± 0.67 [92.35, 94.90] | 93.70 ± 0.96 [91.90, 95.50] | 93.38 ± 1.03 [91.44, 94.91] |
| | Control | 98.93 ± 0.07 [98.8, 99.05] | 95.85 ± 0.13 [95.63, 96.07] | 95.41 ± 0.47 [94.65, 96.16] | 96.25 ± 0.36 [95.71, 96.95] |
| 7.51 | Dysarthria | 98.81 ± 0.14 [98.53, 99.02] | 95.52 ± 0.50 [94.55, 96.41] | 95.56 ± 0.62 [94.35, 96.69] | 95.20 ± 0.72 [93.83, 96.24] |
| | Dysglossia | 99.09 ± 0.06 [98.97, 99.20] | 95.39 ± 0.44 [94.39, 96.20] | 94.98 ± 0.64 [93.53, 96.21] | 97.12 ± 0.57 [96.09, 98.07] |
| | CLP | 97.94 ± 0.13 [97.73, 98.20] | 94.00 ± 0.58 [92.75, 94.91] | 94.32 ± 0.83 [92.52, 95.54] | 92.36 ± 0.94 [90.62, 94.39] |
| | Control | 99.05 ± 0.05 [98.94, 99.14] | 96.16 ± 0.14 [95.90, 96.42] | 96.39 ± 0.47 [95.46, 97.20] | 95.95 ± 0.47 [95.13, 96.71] |

**Supplementary Table 2: Characteristics of the Spanish PC-GITA dataset.** The table reports the total number of speakers and age statistics for different sex subgroups (presented as mean ± standard deviation (SD) and range in years). All recordings were made in controlled noise conditions within a soundproof booth, and the audio was sampled at 44.1 kHz with 16-bit resolution. None of the healthy controls exhibited symptoms of Parkinson's disease or any other neurological disorders.

| Parameter | Overall | Parkinson's Disease | | Healthy | |
| --- | --- | --- | --- | --- | --- |
| | | Female | Male | Female | Male |
| Speakers [n] | 100 | 25 | 25 | 25 | 25 |
| AGE<br>Mean ± SD [years]<br>Range [years] | 61.0 ± 9.3<br>31 - 86 | 60.1 ± 7.8<br>44 - 75 | 62.2 ± 11.2<br>33 - 77 | 60.7 ± 7.7<br>43 - 76 | 61.2 ± 11.3<br>31 - 86 |

**Supplementary Table 3: Detailed evaluation results for Parkinson's Disease detection using the PC-GITA dataset, trained with differential privacy (DP) at different ε values and δ = 0.001.** The table presents the area under the receiver operating characteristic curve (AUROC), accuracy, specificity, and sensitivity, expressed as percentages in the format mean ± standard deviation [95% confidence intervals].

| ε | AUROC | Accuracy | Specificity | Sensitivity |
|---|---|---|---|---|
| ∞ (Non-DP) | 83.27 ± 1.10 [81.41, 85.15] | 81.75 ± 1.35 [79.52, 84.23] | 90.60 ± 2.15 [80.0, 85.97] | 72.90 ± 2.84 [76.25, 83.75] |
| **NO PRETRATING** | | | | |
| 2.51 | 41.53 ± 4.99 [47.47, 59.83] | 51.60 ± 1.75 [55.0, 62.98] | 63.35 ± 40.08 [30.72, 94.44] | 39.85 ± 40.09 [22.78, 86.09] |
| 3.26 | 50.10 ± 4.40 [57.19, 68.49] | 54.68 ± 2.29 [59.52, 66.88] | 58.30 ± 32.11 [34.03, 92.5] | 51.05 ± 32.42 [30.84, 88.75] |
| 4.80 | 65.93 ± 4.34 [66.57, 76.97] | 64.15 ± 3.62 [64.03, 73.12] | 64.90 ± 18.42 [52.5, 83.47] | 63.40 ± 17.22 [50.28, 86.25] |
| 5.46 | 70.44 ± 3.64 [66.81, 77.61] | 67.30 ± 2.90 [63.89, 73.53] | 73.90 ± 11.92 [57.78, 87.5] | 60.70 ± 13.00 [49.03, 81.66] |
| 6.17 | 72.95 ± 3.67 [63.54, 76.25] | 69.60 ± 3.50 [62.5, 74.38] | 66.35 ± 10.61 [49.31, 82.22] | 72.85 ± 11.36 [52.78, 84.16] |
| 7.42 | 73.33 ± 3.87 [67.77, 78.34] | 69.47 ± 3.46 [64.38, 73.75] | 73.45 ± 14.12 [51.53, 86.66] | 65.50 ± 14.12 [54.03, 85.0] |
| 12.59 | 75.80 ± 1.89 [71.75, 76.76] | 73.05 ± 2.14 [68.89, 74.86] | 74.65 ± 6.93 [70.28, 86.25] | 71.45 ± 6.48 [56.81, 72.22] |
| **WITH PRETRATING** | | | | |
| 0.25 | 75.56 ± 1.34 [72.86, 78.23] | 73.55 ± 1.53 [71.25, 76.25] | 72.90 ± 5.46 [62.50, 80.00] | 74.20 ± 5.78 [65.00, 85.00] |
| 1.01 | 75.63 ± 1.26 [73.08, 77.69] | 73.23 ± 1.68 [70.28, 76.25] | 67.65 ± 6.49 [55.56, 77.50] | 78.80 ± 7.22 [65.00, 90.00] |
| 2.18 | 76.68 ± 0.97 [74.99, 78.35] | 75.60 ± 1.55 [73.75, 78.75] | 74.00 ± 6.00 [65.00, 85.00] | 77.20 ± 6.47 [63.06, 86.94] |
| 2.55 | 78.74 ± 0.92 [76.90, 80.35] | 78.03 ± 1.52 [76.25, 81.25] | 75.30 ± 5.14 [67.50, 86.94] | 80.75 ± 5.51 [68.62, 89.44] |
| 4.39 | 80.27 ± 1.06 [78.44, 82.29] | 78.75 ± 1.09 [77.50, 80.00] | 74.70 ± 3.86 [65.00, 80.00] | 82.80 ± 3.70 [77.50, 91.94] |
| 8.86 | 80.34 ± 1.26 [77.68, 82.05] | 76.35 ± 1.43 [73.75, 78.75] | 82.55 ± 6.11 [70.00, 90.00] | 70.15 ± 6.60 [60.00, 82.50] |
| 10.53 | 80.70 ± 1.22 [78.20, 82.60] | 78.03 ± 1.64 [75.00, 80.97] | 87.10 ± 3.92 [80.56, 92.50] | 68.95 ± 4.85 [60.00, 77.50] |

**Supplementary Table 4: Diagnostic performance (AUROC) of private and non-private networks across sex groups.** The results, presented as percentages in the format mean ± standard deviation [95% confidence intervals], report the area under the receiver operating characteristic curve (AUROC) values for Dysarthria, Dysglossia, Cleft Lip and Palate (CLP), the control group, and the overall average across various ε values with δ = 0.001. These metrics are shown separately for the female (n=423) and male (n=437) subgroups of test speakers.

|  |  | ε = 0.87 | ε = 2.77 | ε = 4.29 | ε = 6.36 | ε = 7.51 | ε = ∞ (Non-DP) |
|---|---|---|---|---|---|---|---|
| **Female** | Dysarthria | 94.81 ± 0.35 [94.2, 95.55] | 98.95 ± 0.19 [98.58, 99.22] | 98.93 ± 0.15 [98.66, 99.18] | 99.24 ± 0.12 [99.04, 99.51] | 99.34 ± 0.16 [99.05, 99.55] | 99.98 ± 0.01 [99.94, 100.0] |
|  | Dysglossia | 97.32 ± 0.18 [97.06, 97.66] | 99.20 ± 0.18 [98.85, 99.49] | 99.21 ± 0.12 [99.01, 99.42] | 99.26 ± 0.09 [99.1, 99.42] | 99.37 ± 0.07 [99.26, 99.49] | 99.97 ± 0.01 [99.96, 99.99] |
|  | CLP | 95.47 ± 0.27 [94.96, 95.92] | 97.39 ± 0.26 [96.78, 97.78] | 97.90 ± 0.21 [97.48, 98.25] | 98.74 ± 0.10 [98.56, 98.91] | 98.89 ± 0.12 [98.63, 99.07] | 99.99 ± 0.00 [99.98, 100.0] |
|  | Control | 95.25 ± 0.18 [94.96, 95.57] | 97.61 ± 0.19 [97.32, 97.93] | 98.46 ± 0.13 [98.23, 98.69] | 99.08 ± 0.23 [98.88, 99.21] | 99.19 ± 0.07 [99.07, 99.34] | 99.99 ± 0.01 [99.97, 100.0] |
| **Male** | Dysarthria | 90.46 ± 0.46 [89.45, 91.25] | 97.04 ± 0.28 [96.57, 97.68] | 98.19 ± 0.19 [97.79, 98.52] | 98.09 ± 0.20 [97.82, 98.5] | 98.30 ± 0.23 [97.90, 98.72] | 99.79 ± 0.03 [99.72, 99.84] |
|  | Dysglossia | 95.46 ± 0.18 [95.12, 95.82] | 98.34 ± 0.14 [98.04, 98.63] | 98.66 ± 0.10 [98.46, 98.86] | 98.52 ± 0.12 [98.27, 98.72] | 98.78 ± 0.09 [98.61, 98.94] | 99.93 ± 0.02 [99.89, 99.96] |
|  | CLP | 94.84 ± 0.28 [94.29, 95.3] | 96.52 ± 0.28 [95.98, 96.96] | 97.28 ± 0.24 [96.91, 97.74] | 97.64 ± 0.22 [97.24, 98.0] | 97.49 ± 0.22 [96.99, 97.86] | 99.85 ± 0.05 [99.74, 99.93] |
|  | Control | 96.80 ± 0.13 [96.61, 97.06] | 97.62 ± 0.13 [97.37, 97.86] | 98.60 ± 0.10 [98.43, 98.84] | 98.76 ± 0.11 [98.55, 98.94] | 98.87 ± 0.09 [98.73, 99.03] | 99.84 ± 0.03 [99.78, 99.88] |

**Supplementary Table 5: Performance of private and non-private networks across age groups.** The results, presented as percentages in the format mean ± standard deviation [95% confidence intervals], report the area under the receiver operating characteristic curve (AUROC) values across various ε values with δ = 0.001. Due to an insufficient number of speakers, only Cleft Lip and Palate (CLP) and control groups are analyzed for children and young participants, while only Dysarthria and Dysglossia are analyzed for early adults, middle-aged, and older participants (see **Table 1**). Results are categorized by age groups: children (ages [0, 15), n=417), young participants (ages [15, 30), n=162), early adults (ages [30, 50), n=43), middle-aged (ages [50, 70), n=143), and older participants (ages [70, 100), n=75).

|  |  | ε = 0.87 | ε = 2.77 | ε = 4.29 | ε = 6.36 | ε = 7.51 | ε = ∞ (Non-DP) |
|---|---|---|---|---|---|---|---|
| **[0, 15) years old** | CLP | 83.44 ± 0.68 [81.76, 84.35] | 89.58 ± 0.80 [87.96, 91.07] | 91.10 ± 0.74 [89.70, 92.38] | 93.10 ± 0.54 [92.12, 94.30] | 93.37 ± 0.58 [92.07, 94.21] | 99.70 ± 0.12 [99.43, 99.88] |
|  | Control | 88.87 ± 1.64 [86.38, 91.67] | 95.82 ± 0.30 [95.3, 96.43] | 96.32 ± 0.31 [95.78, 96.81] | 96.75 ± 0.28 [96.24, 97.30] | 96.69 ± 0.33 [95.87, 97.24] | 99.69 ± 0.07 [99.56, 99.81] |
| **[15, 30) years old** | CLP | 94.01 ± 0.53 [93.25, 94.91] | 95.33 ± 0.51 [94.51, 96.32] | 94.79 ± 0.67 [93.30, 95.79] | 94.87 ± 0.50 [93.91, 95.79] | 92.85 ± 0.77 [91.61, 94.22] | 99.70 ± 0.10 [99.51, 99.85] |
|  | Control | 98.28 ± 0.23 [97.85, 98.61] | 98.19 ± 0.38 [97.50, 98.80] | 98.84 ± 0.19 [98.47, 99.17] | 99.11 ± 0.19 [98.74, 99.38] | 99.29 ± 0.11 [99.11, 99.50] | 99.74 ± 0.04 [99.64, 99.81] |
| **[30, 50) years old** | Dysarthria | 79.66 ± 1.26 [77.22, 81.84] | 95.48 ± 0.78 [94.24, 96.87] | 97.54 ± 0.53 [96.61, 98.39] | 97.78 ± 0.53 [96.85, 98.95] | 97.65 ± 0.52 [96.73, 98.58] | 99.99 ± 0.01 [99.97, 100.0] |
|  | Dysglossia | 91.82 ± 0.65 [90.74, 92.99] | 97.66 ± 0.44 [96.87, 98.36] | 98.03 ± 0.34 [97.28, 98.61] | 98.04 ± 0.25 [97.63, 98.48] | 98.13 ± 0.34 [97.58, 98.80] | 99.97 ± 0.02 [99.93, 100.0] |
| **[50, 70) years old** | Dysarthria | 62.06 ± 1.28 [59.46, 64.87] | 85.95 ± 1.28 [83.63, 88.50] | 88.77 ± 0.99 [87.21, 90.73] | 88.05 ± 1.12 [86.12, 89.84] | 88.59 ± 1.05 [86.51, 90.29] | 98.21 ± 0.49 [97.16, 98.93] |
|  | Dysglossia | 79.84 ± 1.88 [75.72, 82.52] | 92.67 ± 0.86 [91.00, 94.25] | 93.19 ± 0.75 [91.98, 94.55] | 93.13 ± 0.81 [91.06, 94.36] | 94.38 ± 0.73 [92.52, 95.49] | 98.90 ± 0.22 [98.43, 99.27] |
| **[70, 100) years old** | Dysarthria | 84.15 ± 1.25 [81.91, 86.51] | 95.22 ± 0.63 [94.06, 96.36] | 96.37 ± 0.48 [95.72, 97.57] | 95.94 ± 0.54 [94.96, 96.93] | 95.64 ± 0.55 [94.58, 96.53] | 99.35 ± 0.24 [98.86, 99.82] |
|  | Dysglossia | 77.39 ± 1.07 [74.82, 79.61] | 93.19 ± 0.85 [91.92, 94.78] | 92.24 ± 0.67 [90.91, 93.60] | 94.23 ± 0.68 [92.99, 95.57] | 93.45 ± 0.84 [91.70, 95.11] | 98.97 ± 0.33 [98.30, 99.61] |

**Example log-Mel-spectrograms of the utilized datasets**

a) Speech disorder dataset

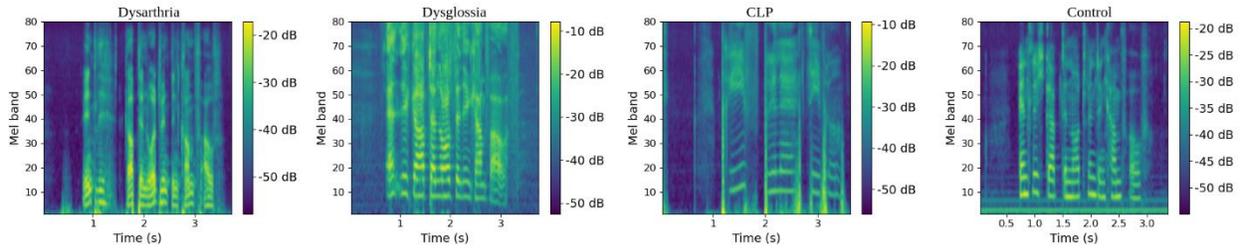

b) PC-GITA

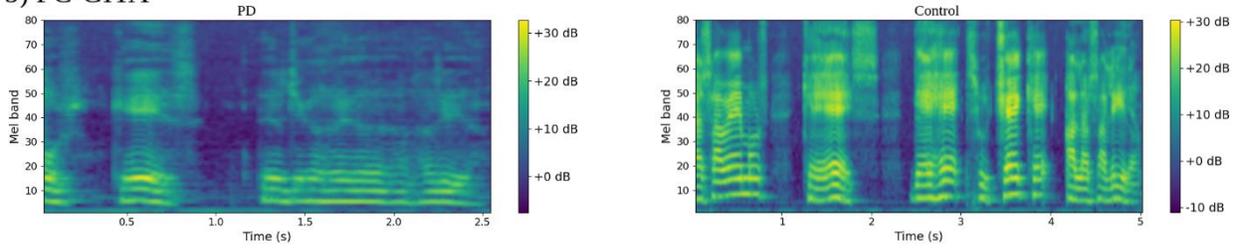

c) LibriSpeech

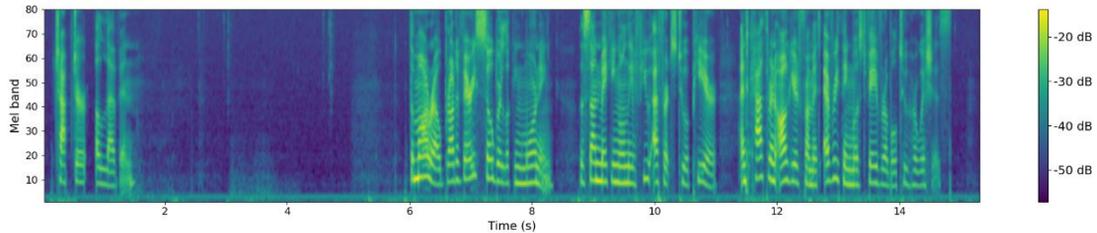

**Supplementary Figure 1: Examples of log-Mel-spectrograms from the speech datasets used in this study.** The figure shows one speech sample from different speakers for each condition, including: **(a)** the German speech disorders dataset with samples for Dysarthria, Dysglossia, Cleft Lip and Palate (CLP), and a healthy control; **(b)** the Spanish PC-GITA dataset with one sample from a Parkinson's Disease (PD) patient and one from a healthy control; and **(c)** a sample from the LibriSpeech dataset of healthy English speakers.